\documentclass[twocolumn,superscriptaddress,amsmath,amssymb,preprintnumbers,aps,prb]{revtex4-1}
\usepackage[colorlinks=true,urlcolor=blue,citecolor=blue,linkcolor=blue]{hyperref}
\usepackage[T1]{fontenc}
\usepackage[latin9]{inputenc}
\usepackage{amssymb}
\usepackage{algpseudocode}
\usepackage{color}
\usepackage{chngcntr}
\usepackage{float}
\usepackage{graphicx}
\usepackage{multirow}
\usepackage{subfigure}
\usepackage{txfonts}
\usepackage{titlesec}
\usepackage{tikz}
\usepackage{ulem}

\def\ket#1{|{#1}\rangle}                 
\def\lsb#1{\left(#1\right)}              
\def\lmb#1{\left[#1\right]}              
\def\llb#1{\left\{#1\right\}}            
\def\Tr#1{\mathrm{Tr}\left(#1\right)}    
\def\op#1{\hat{#1}}                      
\def\kb#1#2{|#1\rangle\langle #2|}       
\def\Softmax{& \multicolumn{4}{c}{Softmax}}
\def\ReLu{& \multicolumn{4}{c}{ReLu}}
\def\wdc#1#2#3{{#1}$\times${#2}$\times${#3}}
\def\BnReLu{& \multicolumn{4}{c}{BN $+$ ReLu}}
\def\tred#1{\textcolor{red}{#1}}

\def\tbrown#1{\textcolor[rgb]{0.5882,0.2941,0}{#1}}

\begin{document}

\title{Compressing deep neural networks by matrix product operators}

\author{Ze-Feng Gao}
\affiliation{Department of Physics, Renmin University of China, Beijing 100872, China}
\author{Song Cheng}
\affiliation{Institute of Physics, Chinese Academy of Sciences, Beijing 100190, China}
\affiliation{Center for Quantum Computing, Peng Cheng Laboratory, Shenzhen 518055, China}
\affiliation{University of Chinese Academy of Sciences, Beijing, 100049, China}
\author{Rong-Qiang He}
\affiliation{Department of Physics, Renmin University of China, Beijing 100872, China}
\author{Z. Y. Xie}
\email{qingtaoxie@ruc.edu.cn}
\affiliation{Department of Physics, Renmin University of China, Beijing 100872, China}
\author{Hui-Hai Zhao}
\email{huihai.zhao@riken.jp}
\affiliation{RIKEN Brain Science Institute, Hirosawa, Wako-shi, Saitama, 351-0106, Japan}
\author{Zhong-Yi Lu}
\email{zlu@ruc.edu.cn}
\affiliation{Department of Physics, Renmin University of China, Beijing 100872, China}
\author{Tao Xiang}
\email{txiang@iphy.ac.cn}
\affiliation{Institute of Physics, Chinese Academy of Sciences, Beijing 100190, China}
\affiliation{University of Chinese Academy of Sciences, Beijing, 100049, China}

\begin{abstract}
  A deep neural network is a parametrization of a multilayer mapping of signals in terms of many alternatively arranged linear and nonlinear transformations.
  The linear transformations, which are generally used in the fully connected as well as convolutional layers, contain most of the variational parameters that are trained and stored.
  Compressing a deep neural network to reduce its number of variational parameters but not its prediction power is an important but challenging problem toward the establishment of an optimized scheme in training efficiently these parameters and in lowering the risk of overfitting.
  Here we show that this problem can be effectively solved by representing linear transformations with matrix product operators (MPOs), which is a tensor network originally proposed in physics to characterize the short-range entanglement in one-dimensional quantum states.
  We have tested this approach in five typical neural networks, including FC2, LeNet-5, VGG, ResNet, and DenseNet on two widely used data sets, namely, MNIST and CIFAR-10, and found that this MPO representation indeed sets up a faithful and efficient mapping between input and output signals, which can keep or even improve the prediction accuracy with a 	dramatically reduced number of parameters. Our method greatly simplifies the representations in deep learning, and opens a possible route toward establishing a framework of modern neural networks which might be simpler and cheaper, but more efficient.
\end{abstract}

\maketitle

\section{Introduction}\label{sec:introduction}

Deep neural networks \cite{DeepCNN, LeNet5, DeepCNN1, NiN, Haiway, DeepCNN2, FracNet, GoogleNet, VGG, ResNet, DenseNet} are important tools of artificial intelligence.
Their applications in many computing tasks, for example, in the famous ImageNet Large Scale Visual Recognition Challenge (ILSVRC) \cite{ILSVRC}, large vocabulary continuous speech recognition \cite{LVCSR}, and natural language processing \cite{NLP}, have achieved great success.
They have become the most popular and dominant machine-learning approaches \cite{DL2015} that are used in almost all recognition and detection tasks \cite{AlexNet, speech}, including but not limited to, language translation \cite{Translation}, sentiment analysis \cite{SentAna}, segmentation and reconstruction \cite{Brain}, drug activity prediction \cite{Drug}, feature identification in big data \cite{DLHE}, and have attracted increasing attention from almost all natural science and engineering communities, including mathematics \cite{XFGu, XFGu2}, physics \cite{DLQPT, MLMelko, NNRG, VAN, Rev}, biology \cite{Brain, DLBio}, and materials science \cite{SC}.

A deep feedforward neural network sets up a mapping between a set of input signals, such as images, and  a set of output signals, say categories, through a multilayer transformation, $\mathcal{F}$, which is represented as a composition of many alternatively arranged linear ($\mathcal{L}$) and nonlinear ($\mathcal{N}$) mappings \cite{NLapp1, NLapp2}.
More specifically, an $n$-layer neural network $\mathcal{F}$ is a sequential product of alternating linear and nonlinear transformations:
\begin{eqnarray}
  \mathcal{F} = \mathcal{N}_n \mathcal{L}_n \cdots \mathcal{N}_{2} \mathcal{L}_{2} \mathcal{N}_{1} \mathcal{L}_{1}.
  \label{GenMap}
\end{eqnarray}
The linear mappings contain most of the variational parameters that need to be determined.
The nonlinear mappings, which contain almost no free parameters, are realized by some operations known as activations, including rectified linear unit, softmax, and so on.

\begin{figure}
\centering
\includegraphics[width=8.5cm]{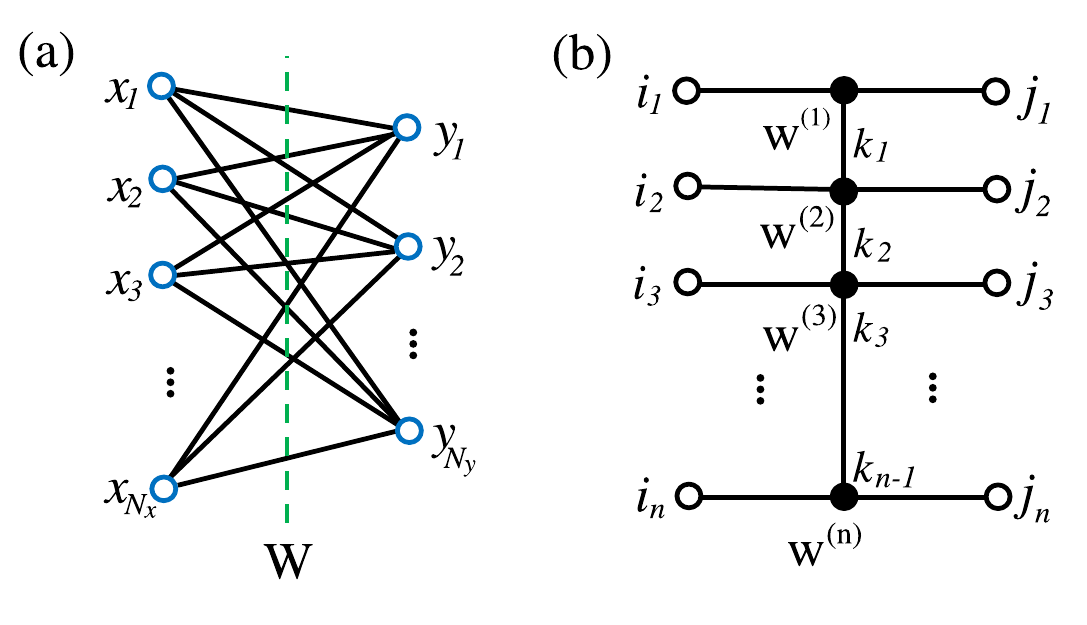}
  \caption{(a) Graphical representation of the weight matrix $W$ in a fully connected layer. The blue circles represent neurons, e.g., pixels.  The solid line connecting an input neuron $x_i$ with output neuron $y_j$ represents the weight element $W_{ji}$.
  (b) MPO factorization of the weight matrix $W$. The local operators $w^{(k)}$ are represented by filled circles.
  The hollow circles denote the input and output indices, $i_l$ and $j_l$, respectively. Given $i_k$ and $j_k$, $w^{(k)}[j_k,i_k]$ is a matrix. }
\label{Fig:MPO}
\end{figure}

A linear layer maps an input vector $\mathbf{x}$ of dimension $N_x$ to an output vector $\mathbf{y}$ of dimension $N_y$ via a linear transformation characterized by a weight matrix $W$:
\begin{eqnarray}
  \mathbf{y} = W \mathbf{x}+\mathbf{b}.   \label{eq:FClayer}
\end{eqnarray}
A fully connected layer plays the role as a global linear transformation, in which each output element is a weighted summation of all input elements, and $W$ is a full matrix.
A convolutional layer \cite{LeNet5} represents a local linear transformation, in the sense that each element in the output is a weighted summation of a small portion of the elements, which form a local cluster, in the input.
The variational weights of this local cluster form a dense convolutional kernel, which is designated to extract some specific features.
To maintain good performance, different kernels are used to extract different features.
A graphical representation of $W$ is shown in Fig. \ref{Fig:MPO}(a).

Usually, the number of elements or neurons, $N_{x}$ and $N_{y}$, are very large, and thus there are a huge number of parameters to be determined in a fully connected layer \cite{VGG}.
The convolutional layer reduces the variational parameters by grouping the input elements into many partially overlapped kernels, and one output element is connected to one kernel.
The number of variational parameters in a convolutional layer is determined by the number of kernels and the size of each kernel.
It could be much less than that in a fully connected layer.
However, the total number of parameters in all the convolutional layers can still be very large in a deep neural network which contains many convolutional layers \cite{ResNet}.
To train and store these parameters raises a big challenge in this field.
First, it is time consuming to train and optimize these parameters, and may even increase the probability of overfitting.
This would limit the generalization power of deep neural networks.
Second, it needs a big memory space to store these parameters.
This would limit its applications where the space of hard disk is strongly confined; for example, on mobile terminals.

There are similar situations in the context of quantum information and condensed-matter physics. In a quantum many-body system, the Hamiltonian or any other physical operator can be expressed as a higher-order tensor in the space spanned by the local basis states \cite{DiracQM}. To represent exactly a quantum many-body system, the total number of parameters that need to be introduced can be extremely huge, and should in principle grow exponentially with the system size (or the size of each ``image'' in the language of neural network). The matrix product operator (MPO) was originally proposed in physics to characterize the short-range entanglement in one-dimensional quantum systems \cite{MPO1, MPO2}, and is now a commonly used approach to represent effectively a higher-order tensor or Hamiltonian with short-range interactions. Mathematically, it is simply a tensor-train approximation \cite{TT, Novikov} that is used to factorize a higher-order tensor into a sequential product of the so-called local tensors. Using the MPO representation, the number of variational parameters needed is greatly reduced since the number of parameters contained in an MPO just grows linearly with the system size. Nevertheless, it turns out that to provide an efficient and faithful representation of the systems with short-range interactions whose entanglement entropies are upper bounded \cite{FrankIllu, AreaLaw} or, equivalently, the systems with finite excitation gaps in the ground states. The application of MPOs in condensed-matter physics and quantum information science has achieved great successes \cite{DMRG, TEBD} in the past decade.

In this paper, we propose to solve the parameter problem in neural networks by employing the MPO representation, which is illustrated in Fig.~\ref{Fig:MPO}(b) and expressed in Eq.~(\ref{Eq:MPO}). The starting point is the observation that the linear transformations in a commonly used deep neural network have a number of similar features as the quantum operators, which may allow us to simplify their representations. In a fully connected layer, for example, it is well known that the rank of the weight matrix is strongly restricted \cite{Compress1, Compress2, Compress3}, due to short-range correlations or entanglements among the input pixels. This suggests that we can safely use a lower-rank matrix to represent this layer without affecting its prediction power. In a convolutional layer, the correlations of images are embedded in the kernels, whose sizes are generally very small in comparison with the whole image size. This implies that the ``extracted features'' from this convolution can be obtained from very local clusters. In both cases, a dense weight matrix is not absolutely necessary to perform a faithful linear transformation. This peculiar feature of linear transformations results from the fact that the information hidden in a data set is just short-range correlated. Thus, to accurately reveal the intrinsic features of a data set, it is sufficient to use a simplified representation that catches more accurately the key features of local correlations.
This motivates us to adopt MPOs to represent linear transformation matrices in deep neural networks.

There have been several applications of tensor network structures in neural networks \cite{Novikov, Garipov, Miles, Kossaifi, Hallam, Liu, MPORNN}. Our approach differs from them by the following aspects: (1) It is physically motivated, emphasizes more on the local structure of the relevant information, and helps to understand the success of deep neural networks. (2) It works in the framework of neural networks, in the sense that the multiple-layer structure and activation functions are still retained, and the parameters are entirely optimized through algorithms developed in neural networks. (3) It is a one-dimensional representation, and is flexible to represent the linear transformations including both the fully-connected layers and the entire convolutional layers. (4) It is also used to characterize the complexity of image datasets; (5) A systematic study has been done. These issues will become clear in the following sections.

The rest of the paper is structured as follows. In Sec.~\ref{sec:Method}, we present the way the linear layers can be represented by MPO and the training algorithm of the resulting network. In Sec.~\ref{sec:Results}, we apply our method systematically to five main neural networks, including FC2, LeNet-5, VGG, ResNet, and DenseNet on two widely used data sets, namely, MNIST and CIFAR-10. Experiments on more data sets can be found in Sec.~\ref{sec:MoreData} in the supplemental materials  \cite{Note1} (SM). Finally, in Sec.~\ref{sec:Discussion}, we discuss the relation with previous efforts, and the possibility to construct a framework of neural networks based on the matrix product representations in the future. In the SM \cite{Note1}, we give the detailed structure of the neural networks used in this work, and provide extra information details about the MPO representations.

\section{Method} \label{sec:Method}
In this paper, the weight matrices $W$ appearing in Eq.~(\ref{eq:FClayer}) and representing linear mappings in the most parameter-consuming layers, to be precise, all the fully connected layers and some of the heaviest convolutional layers  are expressed as MPOs.
To construct the MPO representation of a weight matrix $W$, we first reshape it into a $2n$-indexed tensor:
\begin{equation}
  W_{yx} = W_{j_1j_2\cdots j_n,i_1i_2\cdots i_n} .  \label{Eq:IndexDecompose}
\end{equation}
Here, the one-dimensional coordinate $x$ of the input signal $\mathbf{x}$ with dimension $N_x$ is  reshaped into a coordinate in an $n$-dimensional space, labeled $(i_1 i_2 \cdots i_n)$.
Hence, there is a one-to-one mapping between $x$ and $(i_1 i_2 \cdots i_n)$.
Similarly, the one-dimensional coordinate $y$ of the output signal $\mathbf{y}$ with dimension $N_y$ is also reshaped into a coordinate in an $n$-dimensional space, and there is a one-to-one correspondence between $y$ and $(j_1j_2\cdots j_n)$.
If $I_k$ and $J_k$ are the dimensions of $i_k$ and $j_k$, respectively, then
\begin{equation}
    \prod_{k=1}^{n}I_{k} = N_{x}, \quad \prod_{k=1}^{n}J_{k} = N_{y}  . \label{eq:dims}
\end{equation}
The index decomposition in Eq.~(\ref{Eq:IndexDecompose}) is not unique.
One should in principle decompose the input and output vectors such that the
test accuracy is the highest.
However, to test all possible decompositions is time consuming.
For the results presented in this paper, we have done the decomposition just by convenience, i.e., simply reshaping the single dimension $N_x$ as $n$ parts $\{I_1, I_2,...,I_n\}$ in order. In fact, it can be argued and verified by examples that when the network is away from underfitting, different factorization manners should always produce almost the same result. More details can be found in Sec.~\ref{sec:FacMan} in the SM \cite{Note1}.

The MPO representation of $W$ is obtained by factorizing it into a product of $n$ local tensors,
\begin{eqnarray}
  W_{j_1\cdots j_n,i_1\cdots i_n} = \mathrm{Tr} \left( w^{(1)} [j_1,i_1]w^{(2)} [j_2,i_2] \cdots w^{(n)} [j_n,i_n],  \right) \label{Eq:MPO}
\end{eqnarray}
where $w^{(k)}[j_k,i_k]$ is a $D_{k-1}\times D_{k}$ matrix with $D_k$ the dimension of the bond linking $w^{(k)}$ and $w^{(k+1)}$. In this case, $D_0=D_n=1$.
For convenience in the discussion below, we assume $D_k=D$ for all $k$ except $k=0$ or $n$.
A graphical representation of this MPO is shown in Fig.~\ref{Fig:MPO}(b).

In this MPO representation, the tensor elements of $w^{(k)}$ are variational parameters.
The number of parameters increases with the increase of the bond dimension $D$.
Hence $D$ serves as a tunable parameter that controls the expressive power. In quantum many-body systems, $D$ also controls the expressive accuracy of a target state variationally.

The tensor elements of $w^{(k)}$ in Eq.~(\ref{Eq:MPO}), instead of the elements of $W$ in Eq.~(\ref{eq:FClayer}), are the variational parameters that need to be determined in the training procedure of deep neural networks.
For an MPO whose structure is defined by Eq. (\ref{Eq:M}), the total number of these variational parameters equals
\begin{equation}
   N_{\rm mpo} = \sum_{k=2}^{n-1} I_k J_k D^2 + I_1J_1D + I_nJ_nD ,
\end{equation}
which will be a great reduction of the number $N_xN_y$ in the original fully connected layers (when $N_x$ and $N_y$ are large) and of the number $N_kN_0$ in the original convolutional layers (when the kernel size $N_0$ and the number of kernels $N_k$ are large).

The strategy of training is to find a set of  optimal $w$'s so the following cost function is minimized,
\begin{equation}
L = -\sum_{m}t^{T}_m\log{y_m} + \frac{\alpha}{2}\sum_{i}|w^{(i)}|^2 , \label{cost}
\end{equation}
where $m$ is the label of images, $i$ is the label of all the parameters, including the local tensors in the MPO representations and the kernels in the untouched convolutional layers.
$|w|$ represents the norm of parameter $w$, and $\alpha$ is an empirical parameter that is fixed prior to the training.
The first term measures the cross entropy between prediction vectors $y$ and  target label vectors $t$.
The second term is a constraint, called the L2 regularization \cite{DeepLearn}, which is a widely adopted technique in deep learning to alleviate overfitting, and thus it is also used in all the networks mentioned in this paper, including both the normal neural networks, such as FC2, LeNet-5, VGG, ResNet, as well as DenseNet, and the corresponding MPO-Net counterparts. It should be mentioned that the usage of the L2 regularization has little to do with the validity of the MPO representations, i.e., without L2 regularization, the MPO-Nets should still work as well as the normal networks, as shown in Sec.~\ref{sec:L2} in the SM \cite{Note1}.

To implement a training step, $L$ is evaluated using the known $w$'s, which are randomly initialized, and the input data set.
The gradients of the cost function with respect to the variational parameters are determined by the standard back propagation \cite{BPinDL}.
All the parameters $w$ are treated equally and are updated by the stochastic gradient descent with momentum algorithm \cite{SGDM} in parallel. This is different from the previous effort \cite{Novikov} and is more suitable for deep learning.
This training step is terminated when the cost function stops to drop.

The detailed structures of the neural networks we have studied, as well as the performance on more datasets, different factorization manners, entanglement entropy grasped, the influence of L2 regularization, convergence of training, and so on, are appended systematically in the SM \cite{Note1}. The specific setting of the hyper-parameters for training can be found in the source code \cite{zfgaocode}.
%
\section{Results}\label{sec:Results}

Here we show the results obtained with the MPO representation in five kinds of typical neural networks on two datasets, i.e., FC2 \cite{FC2} and LeNet-5 \cite{LeNet5} on the MNIST data set \cite{MNIST}; VGG \cite{VGG}, ResNet \cite{ResNet}, and DenseNet \cite{DenseNet} on the CIFAR-10 data set \cite{CIFAR}.
Among them, FC2 and LeNet-5 are relatively shallow in the depth of network, while
VGG, Residual CNN (ResNet), and Dense CNN (DenseNet) are deeper neural networks.

For convenience, we use MPO-Net to represent a deep neural network with all or partial linear layers being represented by MPOs.
Moreover, we denote an MPO, defined by Eq.~(\ref{Eq:MPO}), as
\begin{equation}
  M^{J_1,J_2,...,J_n}_{I_1,I_2,...,I_n}(D).
  \label{Eq:M}
\end{equation}
To quantify the compressibility of MPO-net with respect to a neural network, we define its compression ratio $\rho$ as
\begin{equation}
   \rho = \frac {\sum_{l}N^{(l)}_{\text{mpo}}}{\sum_{l} N^{(l)}_{\text{ori}}},
\end{equation}
where $\sum_{l}$ is to sum over the linear layers whose transformation tensors are replaced by MPO.
$N^{(l)}_{\text{ori}}$ and $N^{(l)}_{\text{mpo}}$ are the number of parameters in the $l$-th layer in the original and MPO representations, respectively.
The smaller is the compression ratio, the fewer number of parameters is used in the MPO representation.

Furthermore, to examine the performance of a given neural network, we train the network $m$ times independently to obtain a test accuracy $a$ with a standard deviation $\sigma$ defined by
\begin{eqnarray}
  a &= & \bar{a} \pm \sigma, \\
  \sigma &=& \frac{1}{\sqrt{m-1}} \left[ \sum_{i=1}^{m}(a_i-\bar{a})^2 \right]^{1/2},
\end{eqnarray}
where $a_i$ is the test accuracy of the $i$-th training procedure. $\bar{a}$ is the average of $\{a_i\}$.
The results presented in this paper are obtained with $m=5$.

\subsection{MNIST dataset}

We start from the identification of handwritten digits in the MNIST dataset \cite{MNIST}, which consists of 60000 digits for training and 10000 digits for testing.
Each image is a square of $28 \times 28$ grayscale pixels, and all the images are divided into ten classes corresponding to numbers $0 \sim 9$, respectively.

\subsubsection{FC2}

We first test the MPO representation in the simplest textbook structure of neural network, i.e., FC2 \cite{FC2}. FC2 consists of only two fully connected layers whose weight matrices have $784 \times 256$ and $256 \times 10$ elements, respectively.
We replace these two weight matrices, respectively, by $M_{4,7,7,4}^{4,4,4,4}(D)$ and $M_{4,4,4,4}^{1,1,10,1}(4)$ in the corresponding MPO representation.
Here we fix the bond dimension in the second layer to 4, and only allow the bond dimension to vary in the first layer.

\begin{figure}
\begin{center}
\includegraphics[width=8cm]{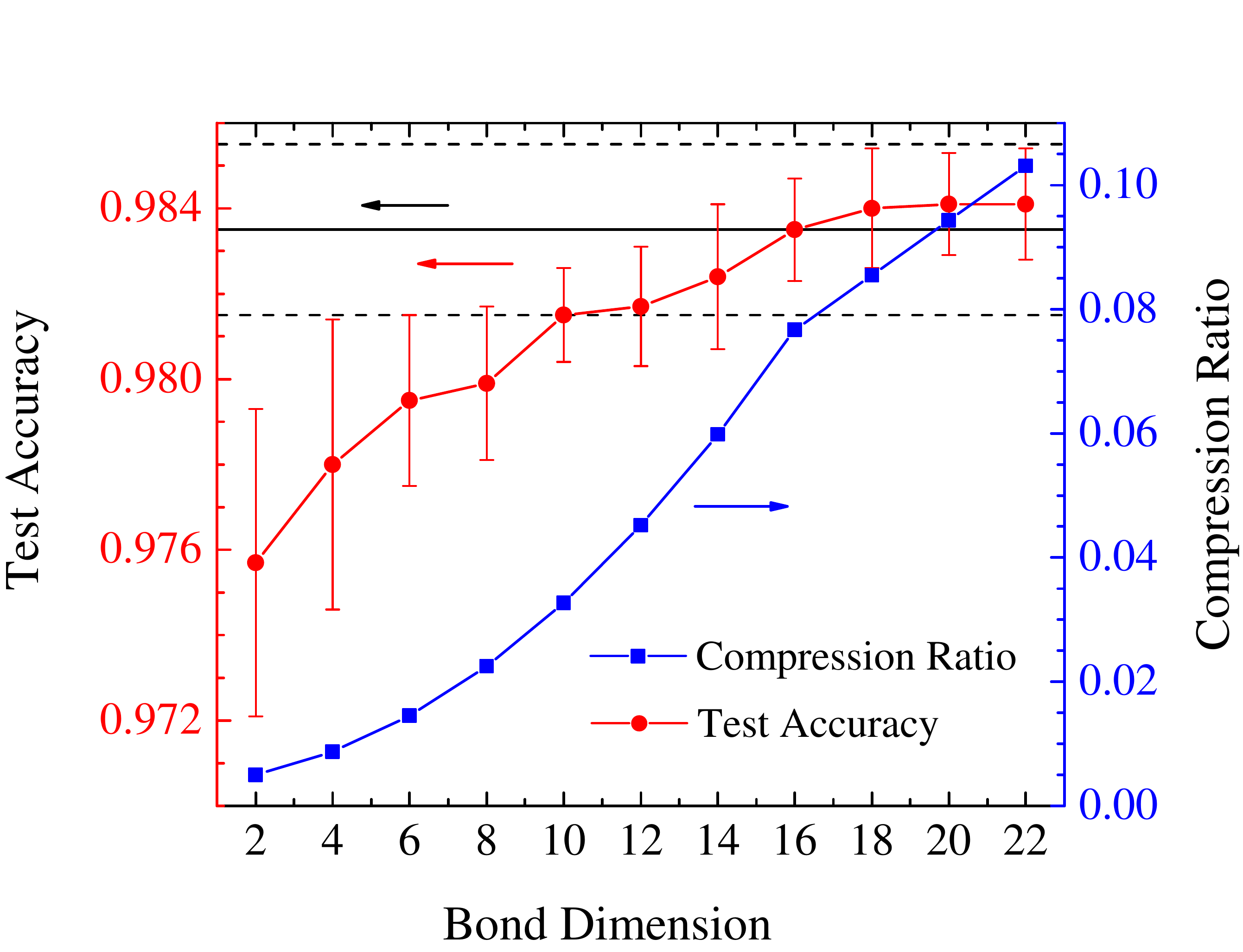}
  \caption{Performance of the MPO representations in FC2 on MNIST. The solid straight line denotes the test accuracy obtained by the normal FC2, $98.35\% \pm 0.2\%$, and the dashed straight lines are plotted to indicate its error bar. }
  \label{fig:MPO-FC2}
\end{center}
\end{figure}

Figure \ref{fig:MPO-FC2} compares the results obtained with FC2 and the corresponding MPO-Net.
The test accuracy of MPO-Net increases when the bond dimension $D$ is increased. It reaches the accuracy of the normal FC2 when $D=16$.
Even for the $D=2$ MPO-Net, which has only 1024 parameters, about 200 times less than the original FC2, the test accuracy is already very good.
This shows that the linear transformations in FC2 are very local and can indeed be effectively represented by MPOs.
The compression ratio of MPO-Net decreases with increasing $D$.
But even for $D=16$, the compression ratio is still below $8\%$, which indicates that the number of parameters to be trained can be significantly reduced without any accuracy loss.

\begin{table}[t]
\centering
  \caption{Test accuracy $a$ and compression ratios $\rho$ obtained in the original and MPO representations  of LeNet-5 on MNIST and VGG on CIFAR-10.}
\label{table:MPO1}
\begin{tabular}{c|c|c|c|c}   \hline\hline
     \multirow{2}*{Data set} &  \multirow{2}*{Network}  & \multirow{2}*{}
      Original Rep & \multicolumn{2}{c}{MPO-Net} \\
      \cline{3-5}
      & & $a$ ($\%$) & $a$ ($\%$) & $\rho$
       \\ \hline 
    MNIST & LeNet-5 & 99.17 $\pm$ 0.04 & 99.17 $\pm$ 0.08 & 0.05 \\ \hline 
  \multirow{2}*{CIFAR-10} &  \multirow{2}*{}  VGG-16       & 93.13$\pm$0.39    &   93.76$\pm$0.16  & $\sim$ 0.0005
   \\   \cline{2-5}
   &   VGG-19     &    93.36$\pm$0.26  & 93.80$\pm$0.09    &    $\sim$0.0005
      \\  \hline \hline
\end{tabular}
\end{table}

\subsubsection{LeNet-5}

We further test MPO-Net with the famous LeNet-5 network \cite{LeNet5}, which is the first instance of convolutional neural networks.
LeNet-5 has five linear layers.
Among them, the last convolutional layer and the two fully connected layers
contain the most of parameters.
We represent these three layers by three MPOs, which are structured as $M_{2,10,10,2}^{2,5,6,2}$(4), $M_{2,5,6,2}^{2,3,7,2}(4)$, and $M_{2,3,7,2}^{1,5,2,1}(2)$, respectively.
The compression ratio is $\rho \sim$ 0.05.

Table \ref{table:MPO1} shows the results obtained with the original and MPO representations of LeNet-5.
We find that the test accuracy of LeNet-5 can be faithfully reproduced by  MPO-Net.
Since LeNet-5 is the first and prototypical convolutional neural network, this success gives us confidence in using the MPO presentation in deeper neural networks.

\subsection{CIFAR-10 dataset}

CIFAR-10 is a more complex dataset \cite{CIFAR}.
It consists of 50000 images for training and 10000 images for testing.  Each image is a square of $32 \times 32$ RGB pixels.
All the images in this data set are divided into ten classes corresponding to airplane, automobile, ship, truck, bird, cat, deer, dog, frog, and horse, respectively.
To have a good classification accuracy, deeper neural networks with many convolutional layers are used.
To show the effectiveness of MPOs representation, as a preliminary test, we use MPO only on the fully connected layers and on some heavily parameter-consuming convolutional layers.

\subsubsection{VGG}

VGG \cite{VGG} is the first \textit{very} deep neural network contrusted.
It won first place in the localization task of the ILSVRC competition 2014.
We have tested two well-established VGG structures, which have 16 and 19 layers, respectively.
In both cases, there are many convolutional layers and three fully connected layers.
We represent the last two heaviest convolutional layers and all the three fully connected layers, respectively, by MPO with the structures: $M^{2,8,8,8,2}_{2,8,8,8,2}(4)$, $M^{2,8,8,8,2}_{2,8,8,8,2}(4)$, $M^{4,4,8,8,4}_{4,4,4,4,2}(4)$, $M^{4,4,8,8,4}_{4,4,8,8,4}(4)$, and $M^{1,10,1,1,1}_{4,4,8,8,4}(4)$.
The result is summarized in Table ~\ref{table:MPO1}.

For both structures, the compression ratio of MPO-Net is about 0.0005. Hence the number of parameters used is much less than in the original representation.
However, we find that the prediction accuracy of MPO-Net is even better than those obtained from the original networks.
This is consistent with the results reported by Novikov \cite{Novikov} for the ImageNet data set \cite{ImagNet}.
It results from two facts of MPOs:
First, since the number of variational parameters is greatly reduced in MPO-Net, the representation is more economical and it is easier to train the parameters.
Second, the local correlations between input and output elements are more accurately represented by MPO.
This can reduce the probability of overfitting.

\subsubsection{ResNet}

\begin{figure}
\centering
\includegraphics[width=8cm]{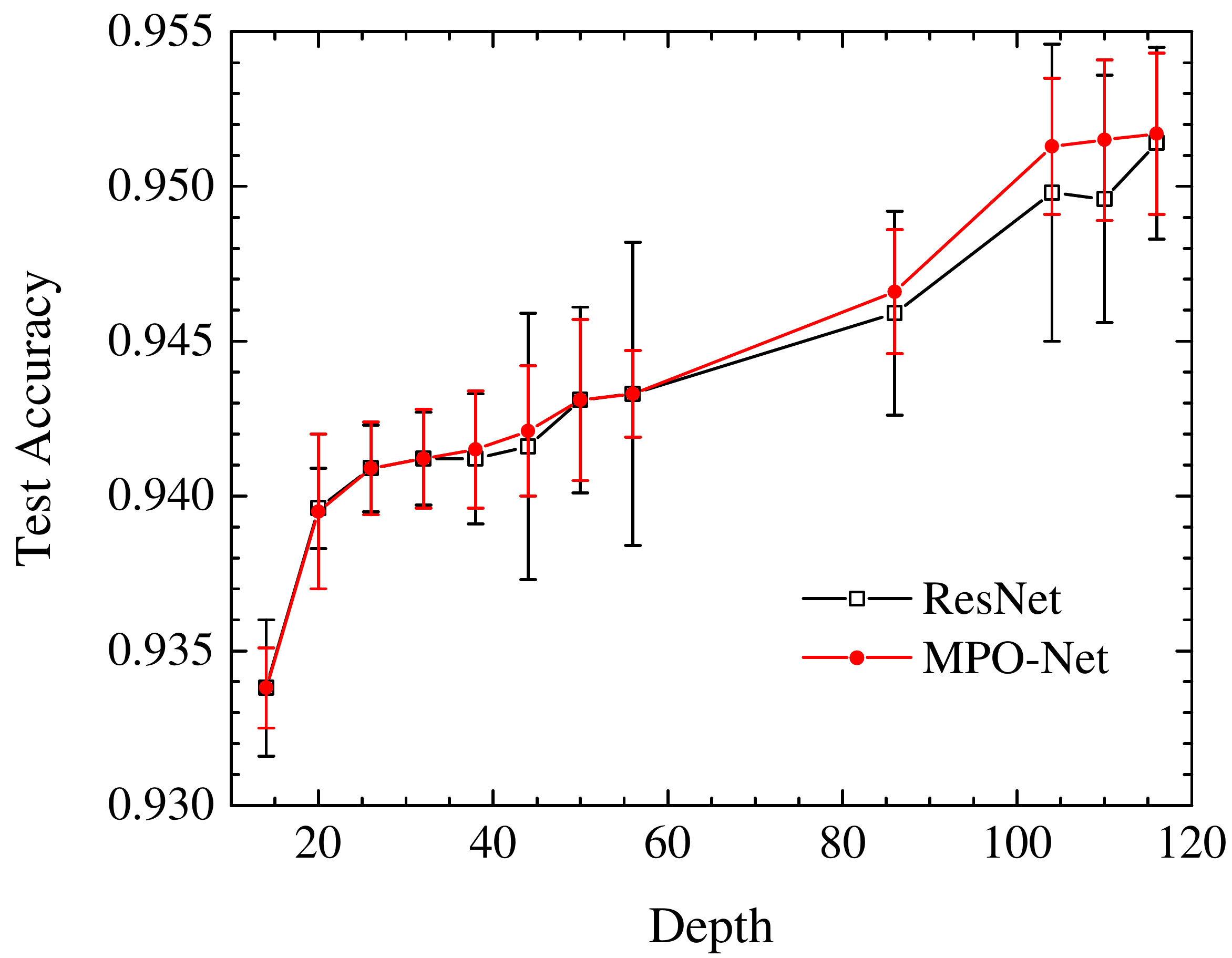}
  \caption{Comparison of the test accuracy $a$ between the original and MPO representations of ResNet on CIFAR-10 with $k=4$. The compression ratio of MPO-Net $\rho \sim 0.11$.}
\label{fig:MPO-ResNet}
\end{figure}

ResNet \cite{ResNet} is commonly used to address the degradation problem with deep convolutional neural networks.
It won first place on the detection task in ILSVRC in 2015, and differs from the ordinary convolutional neural network by the so-called ResUnit structure, in which identity mappings are added to connect some of the input and output signals.
The ResNet structure used in our calculation has a fully connected layer realized by a weight matrix of $64k \times 10$.
Here $k$ controls the width of the network.
We represent this layer by an MPO of $M^{1,5,2,1}_{4,4,4,k}(3)$.
In our calculation, $k = 4$ is use and the corresponding compression ratio is about 0.11.

Figure \ref{fig:MPO-ResNet} shows the test accuracy as a function of the depth of layers of ResNet with $k=4$.
We find that MPO-Net has the same accuracy as the normal ResNet for all the cases we have studied. We also find that even the ResUnit can be compressed by MPO.
For example, for the 56-layer ResNet, by representing the last heaviest ResUnit and the fully connected layer with two $M^{2,4,4,4,4,4,4,2}_{2,4,4,4,4,4,4,2}(4)$ and one $M^{1,5,2,1}_{4,4,4,k}(3)$, we obtain the same accuracy as the normal ResNet.
Similar observations are obtained for other $k$ values.

\subsubsection{DenseNet}

\begin{table*}[htbp]
\centering
\caption{Performance of MPO representations in DenseNet on CIFAR-10.}
\label{tab:MPO-DenseNet}
\begin{tabular}{c|c|c|c|c|c}   \hline\hline
      \multirow{2}*{Depth}       &\multirow{2}*{(n, m, k)} & \multicolumn{2}{c|}{Test accuracy ($\%$)}
             & \multirow{2}*{MPO structure}   & \multirow{2}*{$\rho$} \\
      \cline{3-4} & & DenseNet & MPO-Net & &   \\ \hline
       40         &    (16, 12, 12)         &93.56 $\pm$ 0.26        &93.59 $\pm$ 0.13   & $M^{1,5,2,1}_{4,4,7,4}(4)$  & 0.129      \\  \hline
       40         &    (16, 12, 24)         &95.12 $\pm$ 0.15       &95.13 $\pm$ 0.13    & $M^{1,5,2,1}_{4,5,11,4}(4)$ & 0.089       \\  \hline
       100        &    (24, 32, 12)         &95.36 $\pm$ 0.15       &95.58 $\pm$ 0.07    & $M^{1,5,2,1}_{4,7,7,6}(4)$  & 0.070    \\  \hline
       100        &    (96, 32, 24)         &95.74 $\pm$ 0.09       &96.09 $\pm$ 0.07    & $M^{1,5,2,1}_{5,8,12,5}(4)$ & 0.044     \\  \hline\hline
\end{tabular}
\end{table*}

The last deep neural network we have tested is DenseNet \cite{DenseNet}. Constructed in the framework of ResNet, DenseNet modifies ResUnit to DenseUnit by adding more shortcuts in the units.
This forms a wider neural network, allowing the extracted information to be more efficiently recycled.
It also achieved great success in the ILSVRC competition, and drew much attention in the CVPR conference in 2017.

The DenseNet used in this work has a fully connected layer with a weight matrix of $(n+3km) \times  10$, where $m$ controls the total depth $L$ of the network, $L = 3m+4$, $n$ and $k$ are the other two parameters that specify the network.
There is only one fully connected layer in DenseNet, and we use MPO to reduce the parameter number in this layer.
Corresponding to different $m$, $k$, and $n$, we use different MPO representations.

Our results are summarized in Table \ref{tab:MPO-DenseNet}.
For the four DenseNet structures we have studied, the fully connected layer is compressed by more than seven to 21 times.
The corresponding compression ratios vary from 0.044 to 0.129.
In the first three cases, we find that the test precisions obtained with MPO-Net agree with the DenseNet results within numerical errors.
For the fourth case, the test accuracy obtained with MPO-Net is even slightly higher than that obtained with DenseNet. Further more, in the first structure listed in Table \ref{tab:MPO-DenseNet}, we have also tried to replace the last heaviest convolutional layer by $M^{2,4,4,3,4,2}_{4,5,8,11,4,4}(20)$ while keeping the last fully connected layer represented by $M^{1,5,2,1}_{4,4,7,4}(4)$. The obtained accuracy is about $93.52\pm0.40$, which is still consistent with the original DenseNet result $93.56\pm0.26$. The corresponding compression ratio is also considerable, i.e., $\rho \sim$ 0.497.

Applications to more data sets, such as the Fashion-MNIST \cite{FMNIST} and Street View House Number data sets \cite{SVHN}, can be found in Sec.~\ref{sec:MoreData} in the SM \cite{Note1}. We found that the faithful representation and the effective compression capability of MPO are also valid there.

\section{Discussion}\label{sec:Discussion}

Motivated by the success of MPOs in the study of quantum many-body systems with short-range interactions, we propose to use MPOs to represent linear transformation matrices in deep neural networks.
This is based on the assumption that the correlations between pixels, or the inherent structures of information hidden in ``images'', are essentially localized \cite{ImagEnt, Entropy}, which enables us to make an analogy between an image and a quantum state and further between a linear mapping in neural network and a quantum operator in physics.
We have tested our approach with five different kinds of typical neural networks on two data sets, and found that MPOs can not only improve the efficiency in training and reduce the memory space, as originally expected, but also slightly improve the test accuracy using much fewer number of parameters than in the original networks.
This, as already mentioned, may result from the fact that the variational parameters can be more accurately and efficiently trained due to the dramatic reduction of parameters in MPO-Net.
The MPO representation emphasizes more on the local correlations of input signals.
It puts a strong constraint on the linear transformation matrix and avoids the training data being trapped at certain local minima.
We believe this can reduce the risk of overfitting.

In fact, by using the canonical form \cite{CanoMPS} of an MPO representation obtained from training, we can introduce the entanglement entropy, initially defined for a quantum state in physics, for a dataset in deep learning to quantify the expressive ability of the network and the complexity of the dataset. This can also help to understand the relation between local correlations in the input datasets and the performance of the MPO-Nets. More details about this topic can be found in Sec.~\ref{sec:EE} in the SM \cite{Note1}.

MPOs can be used to represent both fully connected and convolutional layers. In our paper, they are not distinguished from each other at all, and are regarded as the same thing, i.e., linear mappings appeared in Eq.~(\ref{GenMap}).
One can also use it just to represent the kernels in convolutional layers, which is a substantially different approach to use MPOs since the convolutional structure is still retained, as suggested by Garipov \textit{et al.} \cite{Garipov}.
However, it is more efficient in representing a fully connected layer where the weight matrix is a fully dense matrix. This representation can greatly reduce the memory cost and shorten the training time in a deep neural network where all or most of the linear layers are fully connected ones, such as in a recurrent neural network \cite{RNN, MPORNN, SmpoRNN}, which is used to dispose of video data.

Tensor-network representation of deep neural networks is actually not new.
Inspired by the locality assumption about the correlations between pixels, matrix product representation has been already successfully used to characterize and compress images \cite{ImagEnt}, and to determine the underlying generative models \cite{LWPZ}.
Novikov \textit{et al.} \cite{Novikov} also used MPOs to represent some fully connected layers, not including the classifiers, in FC2 and VGG.
Our work, however, demonstrates that all fully connected layers, including the classifiers especially, as well as convolutional layers, can be effectively represented by MPOs no matter how deep a neural network is. In other words,  in our approach there are no concepts of fully connected layers or convolutional layers, but only linear mappings expressed as sparse MPO and parameter-free non-linear activations. We think this is a great simplification for both concepts and applications, and is of great potential due to the much less required memory space and relatively mathematical structure to study. Our work will greatly help the application of neural networks and especially may help to get rid of connection to the cloud.

There are some previous efforts which aim to establish the entire mapping from the input data to the output label, e.g., Stoudenmire and Schwab tried to represent the mapping in terms of a single MPO \cite{Miles}. Our proposal differs from it since we are still working in the framework of neural networks, in the sense that the multiple-layer structure and activation functions are still retained. There are also other mathematical structures that have been used to represent deep neural networks due to entanglement consideration from physics.
For example, Kossaifi \textit{et al.} \cite{Kossaifi} used a Tucker-structure representation, which is a lower-rank approximation of a high-dimensional tensor, to represent a fully connected layer and its input feature.
Hallam \textit{et al.} \cite{Hallam} used a tensor network called a multi-scale entangled renormalization ansatz \cite{MERA} and Liu \textit{et al.} \cite{Liu} used an unitary tree tensor network \cite{TTN} to represent the entire mapping from the input to the output labels. Comparing with these works, our approach is a one-dimensional representation which emphasizes more on local entanglement in physics, and it is more efficient and flexible to represent some intermediate layers.

It is valuable to mention two aspects about the MPO representation. One is about its application scope. Due to the locality assumption, it is expected to work efficiently in the data sets where locality can be appropriately defined; otherwise, a large bond dimension would be necessary, which might lead to the loss of efficiency and advantage. The other is about the manner how the input data is ordered when it is fed to the MPO-Net. It is instructive to notice that different orderings of an input vector are related by elementary transformations; therefore, they should lead to the same prediction, in principle, as long as the bond dimension is sufficiently large; while given a small bond dimension, the ordering which keeps better the locality may lead to higher prediction accuracy. A coarse-grained ordering which can better characterize the locality of the original image was proposed in Ref. [\onlinecite{ImagEnt}]  and is worth being studied systematically in the future.

In this paper, we have proposed to use MPOs to compress the transformation matrices in deep neural networks.
Similar ideas can be used to compress complex data sets, for example, the dataset called ImageNet \cite{ImagNet}, in which each image contains about $224 \times 224$ pixels.
In this case, it is matrix product states \cite{MPS}, instead of MPOs, that should be used.
We believe this can reduce the cost in decoding each ``image'' in a data set, and by combining with the MPO representation of the linear transformation matrices, can further compress deep neural networks and enhance prediction power. A preliminary example is shown in Sec.~\ref{sec:Conv} in the SM \cite{Note1}.
Another possible advance in the future is about the analysis of optimization in a neural network. In this work, in most cases, MPO-Nets converge faster than the original networks in the training procedure, and this is probably due to the far fewer parameters. However, in deep learning, due to the strong nonlinearity of the cost function, e.g., Eq.~(\ref{cost}), more parameters means higher-dimensional variational space and might have more local minima, thus it is difficult for the current optimization approach, e.g., stochastic gradient descent method, to guarantee a faster
convergence speed in a model with fewer parameters. The counter-examples can be found in both normal networks and MPO-Nets, as discussed in Sec.~\ref{sec:Conv} in the SM \cite{Note1}. By using MPO representations, as a complementary tool, we can study this optimization problem in deep learning from the viewpoint of entanglement entropy developed in quantum physics, as we did preliminarily in the SM \cite{Note1}. Therefore, based on these matrix product representations stemming from quantum many-body physics, it is possible to establish a framework of modern neural networks which might be simpler, cheaper, but more efficient and better understood. It is also expected that this bridge between quantum many-body physics and deep learning can eventually provide some useful feedback and insight to physics, and we would like to leave the extensive study as a future pursuit.

\section*{Acknowledgments}

We thank Andrzej Cichocki, Pan Zhang, and Lei Wang for valuable discussions. This work was supported by the National Natural Science Foundation of China (Grants No. 11774420, 11774422, 11874421, 11888101), and by the National R$\&$D Program of China (Grants No. 2016YFA0300503, 2017YFA0302900), and by the Research Funds of Renmin University of China (Grants No. 20XNLG19).

Ze-Feng Gao and Song Cheng contributed equally to this work.


\newpage
\mbox{}
\pagebreak
\widetext
\begin{center}
\textbf{\large Supplemental Materials:\\Compressing deep neural networks by matrix product operators}
\end{center}
\setcounter{equation}{0}
\setcounter{figure}{0}
\setcounter{table}{0}
\setcounter{section}{0}
\setcounter{page}{1}
\setcounter{subsection}{0}

\makeatletter

\def\ket#1{|{#1}\rangle}                 
\def\lsb#1{\left(#1\right)}              
\def\lmb#1{\left[#1\right]}              
\def\llb#1{\left\{#1\right\}}            
\def\Tr#1{\mathrm{Tr}\left(#1\right)}    
\def\op#1{\hat{#1}}                      
\def\kb#1#2{|#1\rangle\langle #2|}       
\def\Softmax{& \multicolumn{5}{c}{Softmax}}
\def\ReLu{& \multicolumn{5}{c}{ReLu}}
\def\wdc#1#2#3{{#1}$\times${#2}$\times${#3}}
\def\BnReLu{& \multicolumn{5}{c}{BN $+$ ReLu}}
\def\tred#1{\textcolor{red}{#1}}
\def\tbrown#1{\textcolor{brown}{#1}}

\titleformat{\section}{\centering\normalfont\bfseries}{Section \thesection.}{1em}{}

\counterwithin*{equation}{section}
\counterwithin*{equation}{subsection}

\maketitle
In this supplemental material, we gives the detailed structure of the neural networks used in this work, and provides extra information about the MPO representation, such as performance on more datasets, different factorization manners, entanglement entropy grasped, the influence of L2 regularization, convergence of training, and so on. These materials will help to understand our work better. The corresponding source code used in this work is available at \href{https://github.com/zfgao66/deeplearning-mpo}{https://github.com/zfgao66/deeplearning-mpo}.

\section{Structure Details of the Neural Networks}\label{sec:Det}
The used structures of FC2, LeNet5, VGG, ResNet, DenseNet in this paper are summarized in Tab.~\ref{FC2}-\ref{DenseNet}, and their prototypes can be found in Ref.~\cite{FC2, LeNet5, VGG, ResNet, DenseNet}, respectively.
In order to simplify the descriptions, firstly we introduce some short-hands summarized in Tab.~\ref{ShortHands} which are used in this materials.

\begin{table}[H]
\centering
\begin{tabular}{c|c} \hline\hline
Abbreviation & Meaning  \\ \hline \hline
MaxPo & a max-pooling layer  \\ \hline
AvgPo & an average-pooling layer \\ \hline
Conv & a convolutional layer  \\ \hline
ConvUnit & a unit composed of convolutional layers \\ \hline
ResUnit & a unit introduced in ResNet  \\ \hline
ResBlock & a block composed of several ResUnits \\ \hline
DenseUnit & a unit introduced in DenseNet \\ \hline
BN & batch normalization  \\  \hline
\multirow{2}*{$\lmb{w,h;c;s}$} & a convolutional layer with $c$ kernels \\
                    &      each with width $w$ height $h$ and stride $s$ \\ \hline
\multirow{2}*{$\lmb{w,h;s}$} & a pooling layer with pooling \\
                    &      width $w$ height $h$ and stride $s$ \\ \hline
\multirow{2}*{$\llb{w,h;c;s,t}$} & a ResUnit composed of two convolutional layers \\
                    &      denoted as width $\llb{w,h;c;s}$ and $\llb{w,h;c;t}$ resp.\\ \hline
$N_{para}$ & number of parameters in the linear layers \\ \hline
\multirow{2}*{Represented} & whether this block of layers are represented \\
                    &      by MPO in the preliminary test \\ \hline
\end{tabular}
\caption{The short-hands used in this materials.}
\label{ShortHands}
\end{table}

\begin{table}[H]
\centering
\begin{tabular}{c|c|c|c|c|c|c}\hline \hline
No. & Layer name & Input size & Output size  & Comment & $N_{para}$ & Represented \\ \hline \hline
1 & FC  & 28$\times$28  &  256  & & 200704 & Yes \\ \hline
  \ReLu \\ \hline
2 & FC  & 256  &  10   & & 2560 & Yes\\ \hline
  \Softmax \\ \hline
\end{tabular}
\caption{The FC2 network structure used in this work. ReLu and Softmax are element-wise operations, whose details are not shown here and after.}
\label{FC2}
\end{table}

\begin{table}[H]
\centering
\begin{tabular}{c|c|c|c|c|c|c}\hline \hline
No. & Layer name & Input size & Output size & Comment & $N_{para}$ & Represented \\ \hline \hline
1 & Conv  & 28$\times$28  &  28$\times$28$\times$6  & [5,5;6;1] & 150 \\  \hline
  \ReLu \\ \hline
  & MaxPo  & 28$\times$28$\times$6  &  14$\times$14$\times$6   & [2,2;2]  \\ \hline
2 & Conv & 14$\times$14$\times$6 & 10$\times$10$\times$16 & $[5,5;16;1]_{np}$ &2400   \\ \hline
  \ReLu  \\ \hline
  & MaxPo  & 10$\times$10$\times$16  &  5$\times$5$\times$16   & [2,2;2] \\ \hline
3 & Conv & 5$\times$5$\times$16 & 120 & $[5,5;120;1]_{np}$  &48000 & Yes\\ \hline
  \ReLu  \\ \hline
4 & FC  & 120  &  84  & & 10080 & Yes \\ \hline
  \ReLu  \\ \hline
5 & FC  & 84  &  10   & & 840 & Yes \\ \hline
  \Softmax  \\ \hline
\end{tabular}
\caption{The LeNet5 network structure used in this work. Here the subscript \emph{np} is used to emphasize that no padding is used in convolutions there.}
\label{LeNet5}
\end{table}

\begin{table}[H]
\centering
\begin{tabular}{c|c|c|c|c|c|c}\hline \hline
No. & Layer name & Input size & Output size & Comment & $N_{para}$ & Represented \\ \hline \hline
1 & ConvUnit  & 32$\times$32$\times$3  &  32$\times$32$\times$64  & 2$\times$[3,3;64;1] & 38592 \\  \hline
  \ReLu \\ \hline
  & MaxPo  & 32$\times$32$\times$64  &  16$\times$16$\times$64   & [2,2;2] \\ \hline
2 & ConvUnit & 16$\times$16$\times$64 & 16$\times$16$\times$128 & 2$\times$[3,3;128;1] & 221184 \\ \hline
  \ReLu  \\ \hline
  & MaxPo  & 16$\times$16$\times$128  &  8$\times$8$\times$128   & [2,2;2] \\ \hline
3 & ConvUnit & 8$\times$8$\times$128 & 8$\times$8$\times$256 & 2$\times$[3,3;256;1]   & 884736 \\ \hline
  \ReLu  \\ \hline
4 & Conv   & \wdc{8}{8}{256}    & \wdc{8}{8}{256} & [1,1;256;1] & 65536 \\ \hline
  \ReLu  \\ \hline
  & MaxPo & \wdc{8}{8}{256}     & \wdc{4}{4}{256} & [2,2;2]     \\ \hline
5 & ConvUnit  & \wdc{4}{4}{256}  &  \wdc{4}{4}{512}  & 2$\times$[3,3;512;1] & 3538944 \\ \hline
  \ReLu  \\ \hline
6  & Conv & \wdc{4}{4}{512}     & \wdc{4}{4}{512} & [1,1;512;1] & 262144 \\ \hline
  \ReLu  \\ \hline
  & MaxPo & \wdc{4}{4}{512}     & \wdc{2}{2}{512} & [2,2;2]     \\ \hline
7 & ConvUnit  & \wdc{2}{2}{512}  & \wdc{2}{2}{512}  & 2$\times$[3,3;512;1] & 4718592 & Yes \\ \hline
  \ReLu  \\ \hline
8 & Conv  & \wdc{2}{2}{512}     & \wdc{2}{2}{512}   & [1,1;512;1] & 262144 \\ \hline
  \ReLu  \\ \hline
  & MaxPo & \wdc{2}{2}{512}     & 512  & [2,2;2]  \\ \hline
9 & FC  & 512  &  4096   &   & 2097152 & Yes \\ \hline
  \ReLu  \\ \hline
10 & FC  & 4096 & 4096    &  & 16777216 & Yes \\ \hline
  \ReLu  \\ \hline
11 & FC  & 4096 & 10      &  & 40960 & Yes \\ \hline
  \Softmax  \\ \hline
\end{tabular}
\caption{The VGG-16 network structure used in this work. Here a ConvUnit denoted with $m\times$ means $m$ convolutional layers separated by ReLu.}
\label{VGG16}
\end{table}

\begin{table}[H]
\centering
\begin{tabular}{c|c|c|c|c|c|c}\hline \hline
No. & Layer name & Input size & Output size & Comment & $N_{para}$ & Represented \\ \hline  \hline
1 & ConvUnit  & 32$\times$32$\times$3  &  32$\times$32$\times$64  & 2$\times$[3,3;64;1] & 38592 \\  \hline
  \ReLu \\ \hline
  & MaxPo  & 32$\times$32$\times$64  &  16$\times$16$\times$64   & [2,2;2] \\ \hline
2 & ConvUnit & 16$\times$16$\times$64 & 16$\times$16$\times$128 & 2$\times$[3,3;128;1] & 221184 \\ \hline
  \ReLu  \\ \hline
  & MaxPo  & 16$\times$16$\times$128  &  8$\times$8$\times$128   & [2,2;2] \\ \hline
3 & ConvUnit & 8$\times$8$\times$128 & 8$\times$8$\times$256 & 4$\times$[3,3;256;1] & 2064384  \\ \hline
  \ReLu  \\ \hline
  & MaxPo & \wdc{8}{8}{256}     & \wdc{4}{4}{256} & [2,2;2]     \\ \hline
4 & ConvUnit  & \wdc{4}{4}{256}  &  \wdc{4}{4}{512}  & 4$\times$[3,3;512;1] & 8257536 \\ \hline
  \ReLu  \\ \hline
  & MaxPo & \wdc{4}{4}{512}     & \wdc{2}{2}{512} & [2,2;2]     \\ \hline
5 & ConvUnit  & \wdc{2}{2}{512}  & \wdc{2}{2}{512}  & 4$\times$[3,3;512;1] & 9437184 & Partially \\ \hline
  \ReLu  \\ \hline
  & MaxPo & \wdc{2}{2}{512}     & 512  & [2,2;2]  \\ \hline
6 & FC  & 512  &  4096   &  & 2097152  & Yes \\ \hline
  \ReLu  \\ \hline
7 & FC  & 4096 & 4096    &  & 16777216 & Yes \\ \hline
  \ReLu  \\ \hline
8 & FC  & 4096 & 10      &  & 40960 & Yes \\ \hline
  \Softmax  \\ \hline
\end{tabular}
\caption{The VGG-19 network structure used in this work. Here a ConvUnit denoted with $m\times$ means $m$ convolutional layers separated by ReLu. By \textit{partially} we mean the last two convolutional layers in that ConvUnit was represented by MPO.}
\label{VGG19}
\end{table}

\begin{table}[H]
\centering
\begin{tabular}{c|c|c|c|c|c|c}\hline \hline
No. & Layer name & Input size & Output size & Comment & $N_{para}$ & Represented \\ \hline \hline
1 & Conv  & 32$\times$32$\times$3  &  32$\times$32$\times$16  & [3,3;16;1] & 432 \\  \hline
  \BnReLu \\ \hline
2 & ResBlock & 32$\times$32$\times$16 & 32$\times$32$\times$16k & m$\times\llb{3,3;16k;1,1}$  & $2304k[(2m-1)k+1]$ \\ \hline
  \BnReLu  \\ \hline
3 & ResUnit  & 32$\times$32$\times$16k  &  16$\times$16$\times$32k   & $\llb{3,3;32k;2,1}$ & $13824k$ \\ \hline
  \BnReLu  \\ \hline
4 & ResBlock & 16$\times$16$\times$32k  & 16$\times$16$\times$32k & (m-1)$\times\llb{3,3;32k;1,1}$ & $18432k^2(m-1)$\\ \hline
  \BnReLu  \\ \hline
5 & ResUnit  & \wdc{16}{16}{32k}  &  \wdc{8}{8}{64k}  & $\llb{3,3;64k;2,1}$ & $55296k$ \\ \hline
  \BnReLu  \\ \hline
5 & ResBlock  & \wdc{8}{8}{64k}  & \wdc{8}{8}{64k}  & (m-1)$\times\llb{3,3;64k;1,1}$ & $73728k^2(m-1)$& Partially \\ \hline
  \BnReLu  \\ \hline
  & AvgPo & \wdc{8}{8}{64k}     & 64k  & [8,8;8]  \\ \hline
6 & FC  & 64k  &  10   &   & 640k & Yes \\ \hline
  \Softmax  \\ \hline
\end{tabular}
\caption{The ResNet network structure used in this work. The total depth $L$ is given by $L = 6m+2$. Here in a single ResUnit, the two convolutional layers are separated by batch normalization and ReLu, and a ResBlock denoted with $m\times$ means $m$ ResUnits separated by batch normalization and ReLu. By \textit{Partially} we mean the last ResUnit in that ResBlock is represented by MPO.}
\label{ResNet}
\end{table}

\begin{table}[H]
\centering
\begin{tabular}{c|c|c|c|c|c|c}\hline \hline
No. & Layer name & Input size & Output size & Comment & $N_{para} $ & Represented \\ \hline  \hline
1 & Conv  & 32$\times$32$\times$3  &  32$\times$32$\times$n  & [3,3;n;1] & $27n$\\  \hline
  \BnReLu \\ \hline
2 & DenseUnit & 32$\times$32$\times$n & 32$\times$32$\times$(n+km) & m$\times$[3,3;k;1] & $9km[n+k(m-1)/2]$ \\ \hline
  \BnReLu  \\ \hline
3 & Conv  & 32$\times$32$\times$(n+km) & 32$\times$32$\times$(n+km) & [1,1;n+km;1] & $(n+km)^2$ \\ \hline
  & AvgPo & \wdc{32}{32}{(n+km)}   & \wdc{16}{16}{(n+km)}  & [2,2;2]  \\  \hline
  \BnReLu  \\  \hline
4 & DenseUnit  & 16$\times$16$\times$(n+km)  &  16$\times$16$\times$(n+2km)   & m$\times$[3,3;k;1] & $9km[n+km+k(m-1)/2]$   \\ \hline
  \BnReLu  \\ \hline
5 & Conv & 16$\times$16$\times$(n+2km)  & 16$\times$16$\times$(n+2km) & [1,1;n+2km;1] & $(n+2km)^2$ \\ \hline
  & AvgPo  &  \wdc{16}{16}{(n+2km)}   & \wdc{8}{8}{(n+2km)}  & [2,2;2]  \\   \hline
  \BnReLu  \\ \hline
6 & DenseUnit  & \wdc{8}{8}{(n+2km)}  &  \wdc{8}{8}{(n+3km)}  & m$\times$[3,3;k;1] & $9km[n+2km+k(m-1)/2]$ & Partially \\ \hline
  \BnReLu  \\ \hline
  & AvgPo & \wdc{8}{8}{(n+3km)}     & n+3km  & [8,8;8]  \\ \hline
7 & FC  & n+3km  &  10   &  & 10(n+3km) & Yes \\ \hline
  \Softmax  \\ \hline
\end{tabular}
\caption{The DenseNet network structure used in this work. The total depth $L$ is given by $L = 3m+4$. A DenseUnit denoted with $m\times$ means there are $m$ convolutional layers separated by batch normalization and ReLu in this unit. By \textit{Partially} we mean the last convolutional layer there is represented by MPO.}
\label{DenseNet}
\end{table}

\section{Extra information about MPO-Net}
\subsection{Performance on more datasets}\label{sec:MoreData}
In order to show the validity of the MPO representation further, we have also applied it to another two more sophisticated datasets than MNIST, i.e., the Fashion-MNIST \cite{FMNIST} and the Street View House Number (SVHN) \cite{SVHN}.

\begin{figure}
\centering
\includegraphics[width=8cm]{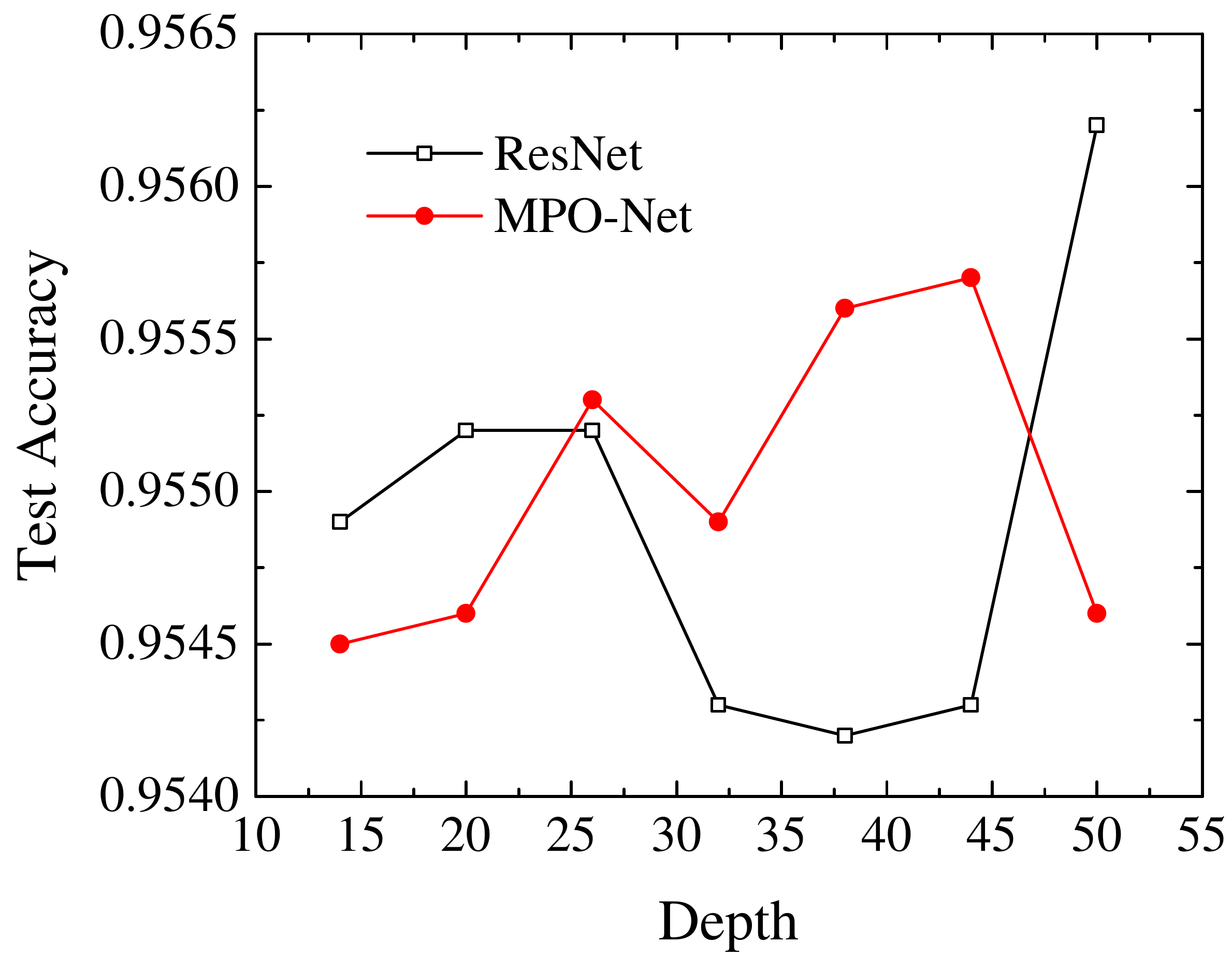}
  \caption{Comparison of the test accuracy $a$ between the original and MPO representations of ResNet on Fashion-MNIST with $k=4$. The compression ratio of MPO-Net $\rho\sim$ 0.11. \textbf{Note}: in this plot, the standard deviations ($\sigma$) of ResNet and MPO-Net are about 0.002 and 0.001, respectively, therefore, the two lines can be regarded as very consistent at all depthes.}
\label{Fig:Fmnist}
\end{figure}

The Fashion-MNIST consists of a training set of 60,000 examples and a test set of 10,000 examples. Each image is a square of $28\times 28$ grayscale pixels, and all the images are divided into 10 classes corresponding to T-shirt, trousers, pullover, dress, coat, sandal, shirt, sneaker, bag, and ankle boot, respectively. We have tested MPO representations in VGG-16, VGG-19, and the ResNet. The detailed structures are already listed in Tab. \ref{VGG16}, \ref{VGG19}, and \ref{ResNet}. As shown in Fig.~\ref{Fig:Fmnist}, with depth up to 50, the MPO can indeed faithfully represent the linear transformations in the ResNet. And the performance in VGG network is shown in Tab.~\ref{MoreData}, which also evidences the validity of MPO in both VGG-16 and VGG-19.

\begin{table}[t]
\centering
\begin{tabular}{c|c|c|c|c}   \hline\hline

     \multirow{2}*{Dataset} &  \multirow{2}*{Network}  & Original Rep & \multicolumn{2}{c}{MPO-Net} \\
      \cline{3-5}
      & & $a$ ($\%$) & $a$ ($\%$) & $\rho$
       \\ \hline \hline
  \multirow{2}*{Fashion-MNIST} &  \multirow{2}*{}  VGG-16       & 94.96    &   94.88  & $\sim$0.0005
   \\   \cline{2-5}
   &   VGG-19     &    95.07  & 94.95    &    $\sim$0.0005
      \\  \hline \hline
  \multirow{3}*{SVHN} &  \multirow{3}*{}  VGG-16       & 96.166    &   96.331  & $\sim$0.0005
   \\   \cline{2-5}
   &   VGG-19     &    96.266  & 96.412    &    $\sim$0.0005
   \\   \cline{2-5}
   &   DenseNet   &    96.354 &  96.235    &    $\sim$0.129
      \\  \hline \hline
\end{tabular}
\caption{Test accuracy $a$ and compression ratios $\rho$ obtained in the original and MPO representations in VGG and DenseNet. \textbf{Note}: in this table, the standard deviations ($\sigma$) of VGG16, VGG19 and DenseNet are about 0.4\%, 0.2\%, 0.2\%, respectively, while the standard deviations ($\sigma$) of all relative MPO-nets are about 0.1\%. Therefore, the results from normal networks and the MPO counterparts are consistent essentially.}
\label{MoreData}
\end{table}

\begin{figure}
\centering
\includegraphics[width=8cm]{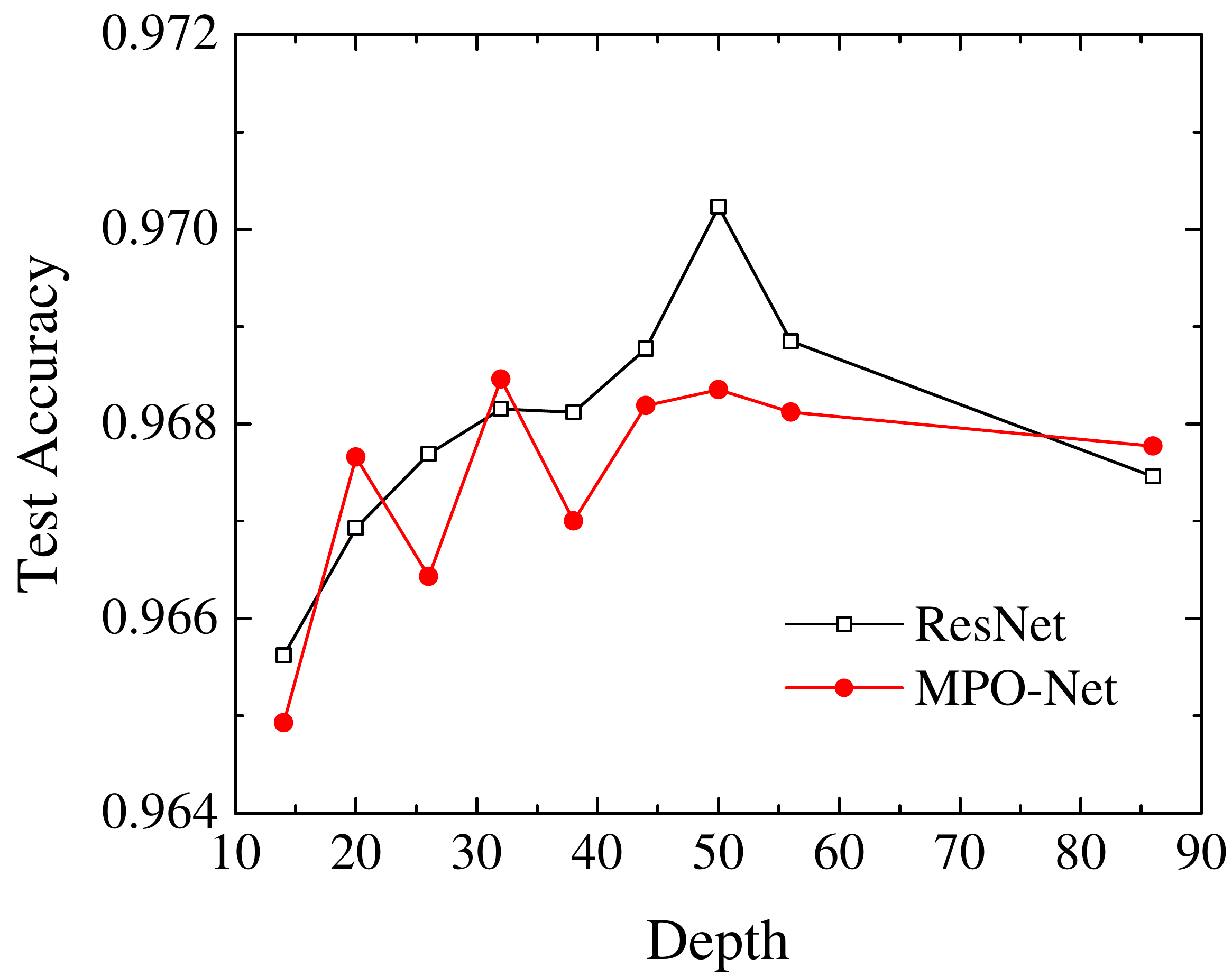}
  \caption{Comparison of the test accuracy $a$ between the original and the MPO representations of ResNet on SVHN with $k=4$. The compression ratio of MPO-Net $\rho\sim$ 0.11. \textbf{Note}: in this plot, the standard deviations ($\sigma$) of ResNet and MPO-Net are about 0.002 and 0.001, respectively, therefore, the two lines can be regarded as very consistent at all depthes.}
\label{Fig:SVHN}
\end{figure}

For the SVHN dataset, it consists of a training set of 73,257 examples and a test set of 26,032 examples. Each image is a square of $32\times 32$ RGB pixels coming from significantly harder, unsolved, real world problem (recognizing digits and numbers in natural scene images), and all the images are divided into 10 classes corresponding to numbers 0$\sim$9, respectively. We have tested MPO representations in the VGG-16, VGG-19, ResNet, and the DenseNet. The detailed structures are listed in Tab.~\ref{VGG16}, \ref{VGG19}, \ref{ResNet}, and \ref{DenseNet}. The results are summarized in Fig.~\ref{Fig:SVHN} and Tab.~\ref{MoreData}. The conclusion is the same as in the case of Fashion-MNIST, i.e., MPO can work well in these networks.

It is definitely impossible to exhaust all the existing datasets and neural networks, but the validity in the five networks, i.e., FC2, LeNet-5, VGG-16(19), ResNet, and DenseNet, on the four datasets, i.e., MNIST, CIFAR-10, Fashion-MNIST, and SVHN, should have provided a strong evidence that MPO can indeed represent the linear transformations faithfully, especially the fully-connected layers and heaviest convolutional layers, in a general deep neural networks.

\subsection{Factorization manner}\label{sec:FacMan}
As claimed in the main text, the factorization manner of the hidden neurons, i.e., how the $N_x$ and $N_y$ are decomposed into $I_k$ and $J_k$ in Eq.(4), is chosen by convenience due to fact that testing all possible decompositions is very time-consuming. In the viewpoint of optimization, different factorization manners affect directly the number of variational parameters, thus will definitely affect the test accuracy. However, in the viewpoint of entanglement entropy, that is controlled by the bond dimension $D$, as $D$ becomes larger, the entanglement entropy of the input data grasped by MPO grows accordingly and the accuracy is expected to be improved, thus $D$ will be more important as long as the network is away from underfitting. In other word, when $D$ becomes larger, different factorization manner should not make too much differences.

\begin{figure}
\centering
\includegraphics[width=9cm]{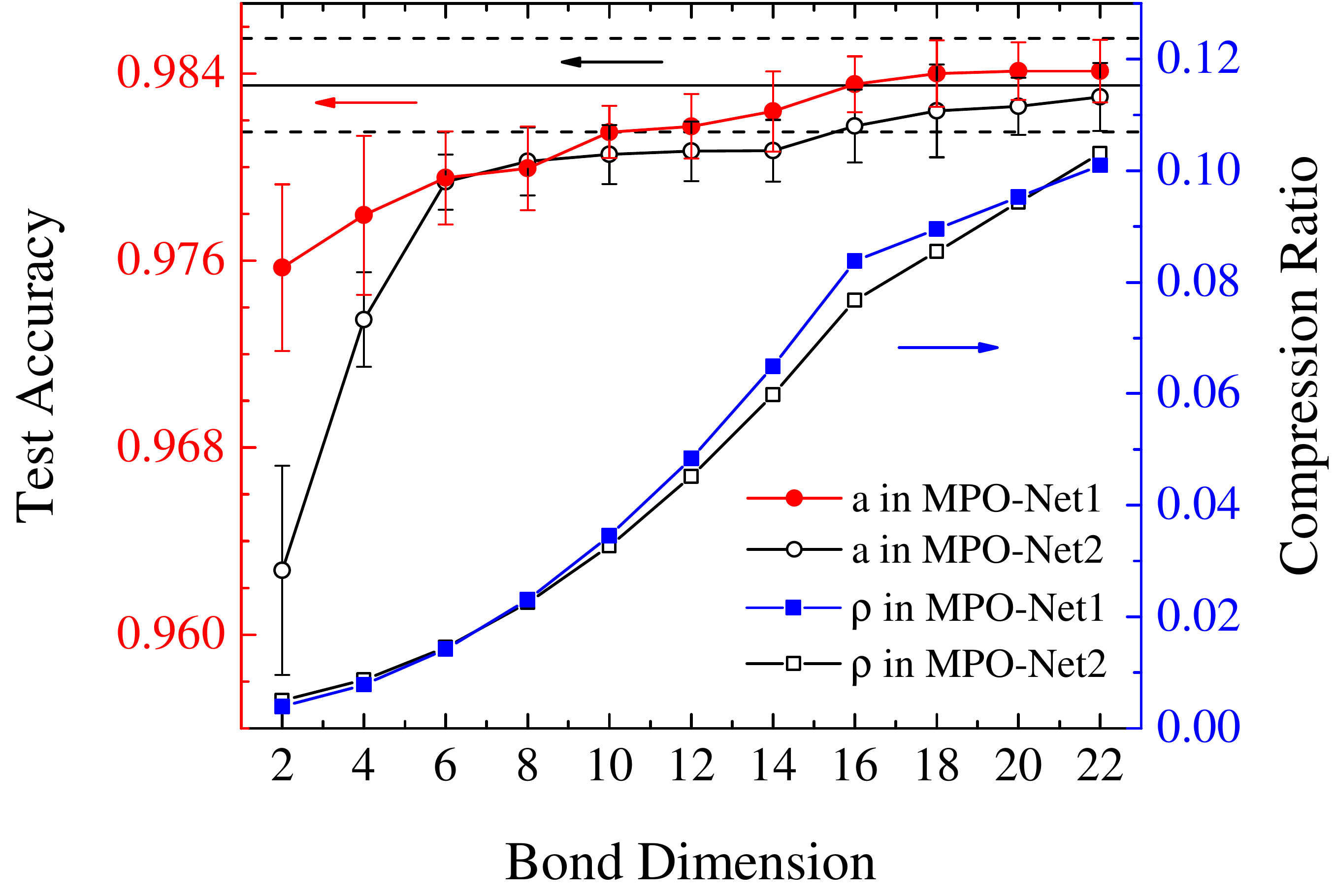}
  \caption{Comparison of the test accuracy on MNIST dataset between two different MPO-Nets. The two MPO-Nets are differed by using different factorization manners of the fully-connected layers in the same FC2 structure. The solid straight line denotes the test accuracy obtained by the normal FC2, $98.35\% \pm 0.2\%$, and the dashed straight lines are also plotted to indicate its error bar.}
\label{Fig:Factor}
\end{figure}

We verified this argument in two experiments. One is the FC2 network on MNIST dataset, the other is the VGG-19 network on CIFAR-10 dataset. In both cases, MPO-Net1 is the network we used in the main text, whose details is listed either in Tab.~\ref{FC2} and Tab.~\ref{VGG19}.
As to the FC2 network, the result is shown in Fig.~\ref{Fig:Factor}, where MPO-Net2 uses $M^{2,2,4,4,2,2}_{2,2,7,7,2,2}(D)$ and $M^{1,1,2,5,1,1}_{2,2,4,4,2,2}(D)$ to represent the two fully-connected layers there. It shows clearly that when $D\geq 4$, the two MPO-Nets produce the same test accuracy essentially, and both of them can reproduce the FC2 result. For the VGG-19 network, the result is shown in Tab.~\ref{Tab:Factor}, where MPO-Net2 uses $M^{2,8,8,8,2}_{2,8,8,8,2}(4)$, $M^{2,8,8,8,2}_{2,8,8,8,2}(4)$, $M^{2,8,8,8,4}_{2,4,8,4,2}(4)$, $M^{2,8,8,8,4}_{2,8,8,8,4}(4)$ and $M^{1,10,1,1,1}_{2,8,8,8,4}(4)$ to represent the last five heaviest linear transformations. The obtained accuracy of the two MPO-Nets are identical within the error bar, and both of them coincide with, and a little better than, the normal VGG-19 result, $93.36\%\pm0.09\%$.

In both cases, the obtained results support our argument above, i.e., when $D$ becomes larger, different factorization manners produce almost the same result.

\begin{table}[t]
\centering
\begin{tabular}{c|c|c|c|c|c}   \hline\hline
     \multirow{2}*{Dataset} &  \multirow{2}*{Network}  & \multicolumn{2}{c|}{MPO-Net1} & \multicolumn{2}{c}{MPO-Net2}  \\
      \cline{3-6}
      & & $a$ ($\%$) & $\rho$ & $a$ ($\%$) & $\rho$
       \\ \hline \hline
    CIFAR-10 & VGG-19 & 93.80 $\pm$0.09 & 0.0005 & 93.78 & 0.0006 \\ \hline \hline
\end{tabular}
\caption{Comparison of the test accuracy on CIFAR-10 dataset between two different MPO-Nets. The two MPO-Nets are differed by using different factorization manners of the three fully-connected layers in the same VGG-19 structure. Due to limited resources, MPO-Net2 has been run 1 time only, but the result is already consistent with that obtained from MPO-Net1.}
\label{Tab:Factor}
\end{table}

\subsection{Entanglement entropy}\label{sec:EE}
Since entanglement entropy is only defined for a quantum state, there is no widely-accepted approach to measure the entanglement entropy of a set of classical configurations, e.g., image pixels in a dataset. Nevertheless, in MPO-Nets, it is possible to use the entanglement spectra of the obtained MPO from training to define an entanglement entropy, which can be regarded as an approximated characterization of the entanglement entropy hidden in the underlying input dataset. Since in quantum physics, entanglement entropy is obtained by bipartition of the full system, it can be argued that the larger the entanglement entropy is, the less local the entanglement or correlation in the dataset is, and vice versa. In this section, we use this approach to compare the entanglement entropy between MNIST and Fashion-MNIST.

\begin{figure}
\centering
\begin{tikzpicture}
\draw[very thick] (0, 0) -- (1.5, 0) -- (3, 0) -- (4.5, 0);
\draw[very thick] (0, 0) -- (0, -1);
\draw[very thick] (0, 0) -- (0, 1);
\draw[very thick] (1.5, 0) -- (1.5, -1);
\draw[very thick] (1.5, 0) -- (1.5, 1);
\draw[very thick] (3, 0) -- (3, -1);
\draw[very thick] (3, 0) -- (3, 1);
\draw[very thick] (4.5, 0) -- (4.5, -1);
\draw[very thick] (4.5, 0) -- (4.5, 1);
\draw (0.75, 0.2) node(X) {bond1};
\draw (2.25, 0.2) node(X) {bond2};
\draw (3.75, 0.2) node(X) {bond3};
\draw (0, 1.25) node(X) {$j_1$};
\draw (0, -1.25) node(X) {$i_1$};
\draw (1.5, 1.25) node(X) {$j_2$};
\draw (1.5, -1.25) node(X) {$i_2$};
\draw (3, 1.25) node(X) {$j_3$};
\draw (3, -1.25) node(X) {$i_3$};
\draw (4.5, 1.25) node(X) {$j_4$};
\draw (4.5, -1.25) node(X) {$i_4$};
\end{tikzpicture}
\caption{MPO configuration, $M_{4,7,7,4}^{4,4,4,4}$, used in the calculation of entanglement entropy. As listed in Tab.~\ref{FC2} and in the main text, all the dimensions $I$s and $J$s are 4, except $I_2 = I_3 = 7$ due to the input size is 28 $\times$ 28.}
\label{Fig:MPOconfig}
\end{figure}
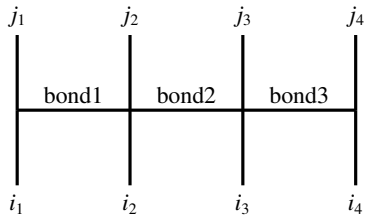

To be specific, to simplify the idea we train the MPO-Net counterpart of the FC2 network on the two datasets respectively, and then use the first MPO in the networks to obtain the entropy. In our calculation, the first MPO in both cases, $M_{4,7,7,4}^{4,4,4,4}$, can be illustrated as in Fig.~\ref{Fig:MPOconfig}. After training, all the four local tensors are known, and we can obtain the normalized entanglement spectra by either obtaining the canonical form \cite{CanoMPS} of the MPO representation or decomposing the corresponding matrix through singular value decomposition. E.g., for bond1 in Fig.~\ref{Fig:MPOconfig}, we can fuse the indices to form a matrix $T_{i_1j_1,i_2i_3i_4j_2j_3j_4}$ and obtain its singular values $v_i$, then the entanglement spectra $\lambda$ and entanglement entropy $S$ can be obtained in the following
\begin{equation}
S = -\sum_{i}\lambda_i\ln\lambda_i, \quad \text{where} \quad \lambda_i = \frac{v_i^2}{\sum{v_i^2}}
\end{equation}
The entropy measured from bond2 and bond3 can be obtained similarly.

\begin{figure}
\centering
\includegraphics[width=7.8cm]{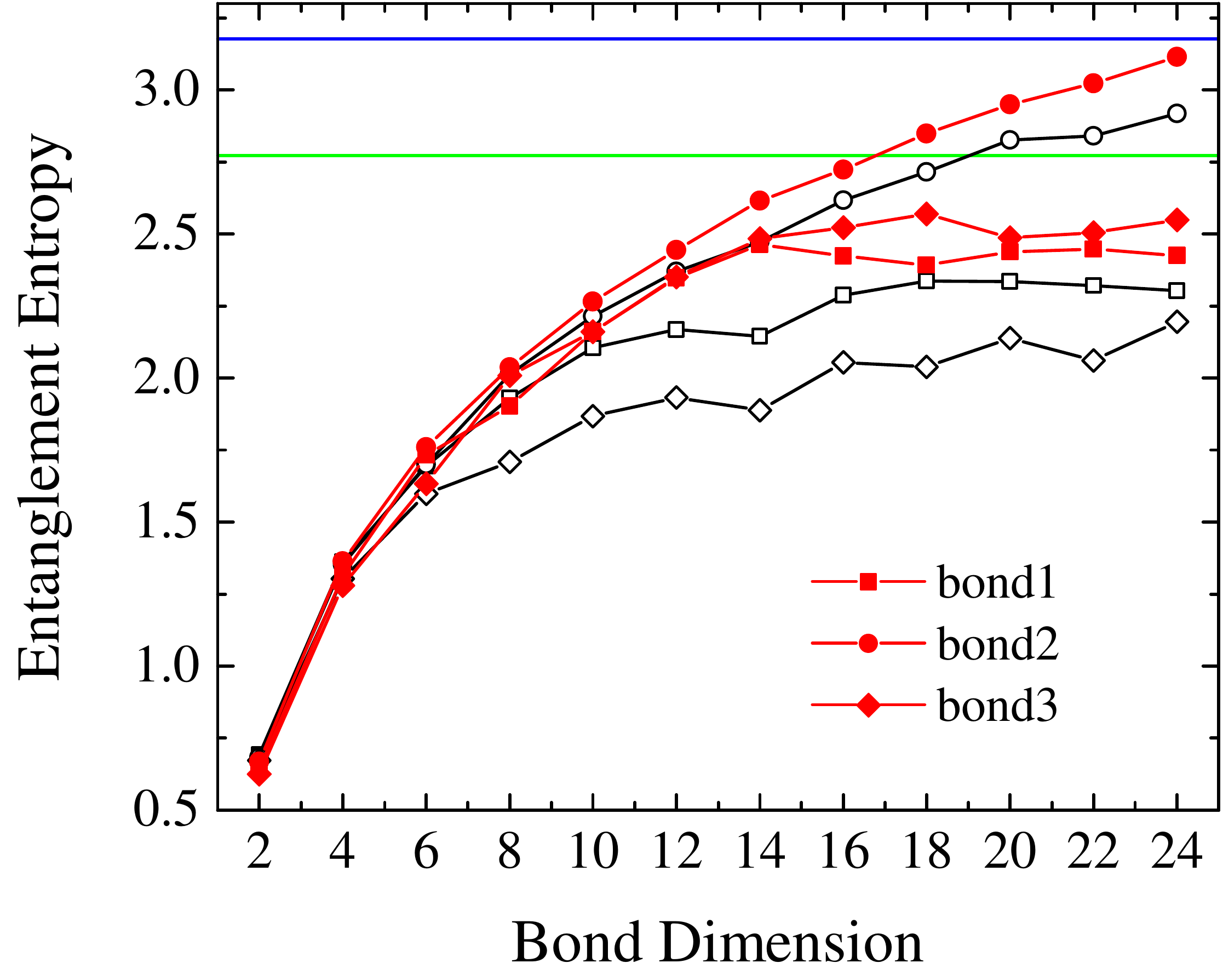}
  \caption{Entanglement entropy measured from three bonds as shown in Fig.~\ref{Fig:MPOconfig} and obtained from the first MPO in the two FC2-MPO networks after well-trained on the MNIST (black curves) and Fashion-MNIST (red curves) datasets. The two straight lines are the theoretical upper bounds of the entropies that the MPO-Net could grasp, i.e., blue line is $\ln 24$ for bond2, and green line is $\ln 16$ for bond1 and bond3.}
\label{Fig:Entropy}
\end{figure}

The result is summarized in Fig.~\ref{Fig:Entropy}. It shows that for all the three bonds, the entanglement entropy obtained from Fashion-MNIST is generally larger than that obtained from MNIST, especially the entropy on bond2 from Fashion-MNIST almost saturates to the upper bound when $D=24$. This probably means that the correlation in MNIST is much more local than Fashion-MNIST, which is consistent with the fact that FC2 and its MPO-Net counterpart can work much better in MNIST, while for Fashion-MNIST, to obtain a comparative accuracy, one has to use more complicated network, e.g., VGG and ResNet, as shown in Fig.~\ref{Fig:Fmnist} and Tab.~\ref{MoreData}, or their MPO-Net counterparts. This example shows that MPO-Net, as well as its normal counterpart, indeed perform better in the systems where local entanglement or correlation plays an important role, while for the systems where most of the correlation is long-range, they probably perform worse. In fact, in our construction, for the Fashion-MNIST data, FC2 and LeNet-5, and their MPO-Net counterparts, can only give a test accuracy about 90$\%$, much lower than that in MNIST.

By the way, it seems also in Fig.~\ref{Fig:Entropy} that the entropy converges to a rather lower value than the theoretical limit on the 1st and 3rd bond for both two datasets, and this actually addresses the existence of locality in the MNIST and Fashion-MNIST datasets, and provides the possibility to apply MPO-Nets on these datasets (but with more layers probably).

\subsection{About L2 regularization}\label{sec:L2}
Essentially, L2 regularization can be regarded as a constrain on the variational parameters in the neural networks. In the field of deep learning, it is believed to be able to alleviate overfitting, and is widely used in modern neural networks \cite{DeepLearn}. Therefore, in all the networks mentioned in this work, including both the normal neural networks, such as FC2, LeNet-5, VGG, ResNet, and DenseNet, as well as the corresponding MPO-Net counterparts, the L2 regularization is always used. However, the validity of MPO representation does not rely on the usage of regularization, i.e., without L2 regularization, MPO-Net should still works well. In this section, we redo the comparison in Fig.2 and Table.1 in the main text, i.e., performance comparison between MPO-Net and normal networks on MNIST,  to verify this statement, but without L2 regularization.

\begin{figure}
\centering
\includegraphics[width=8.5cm]{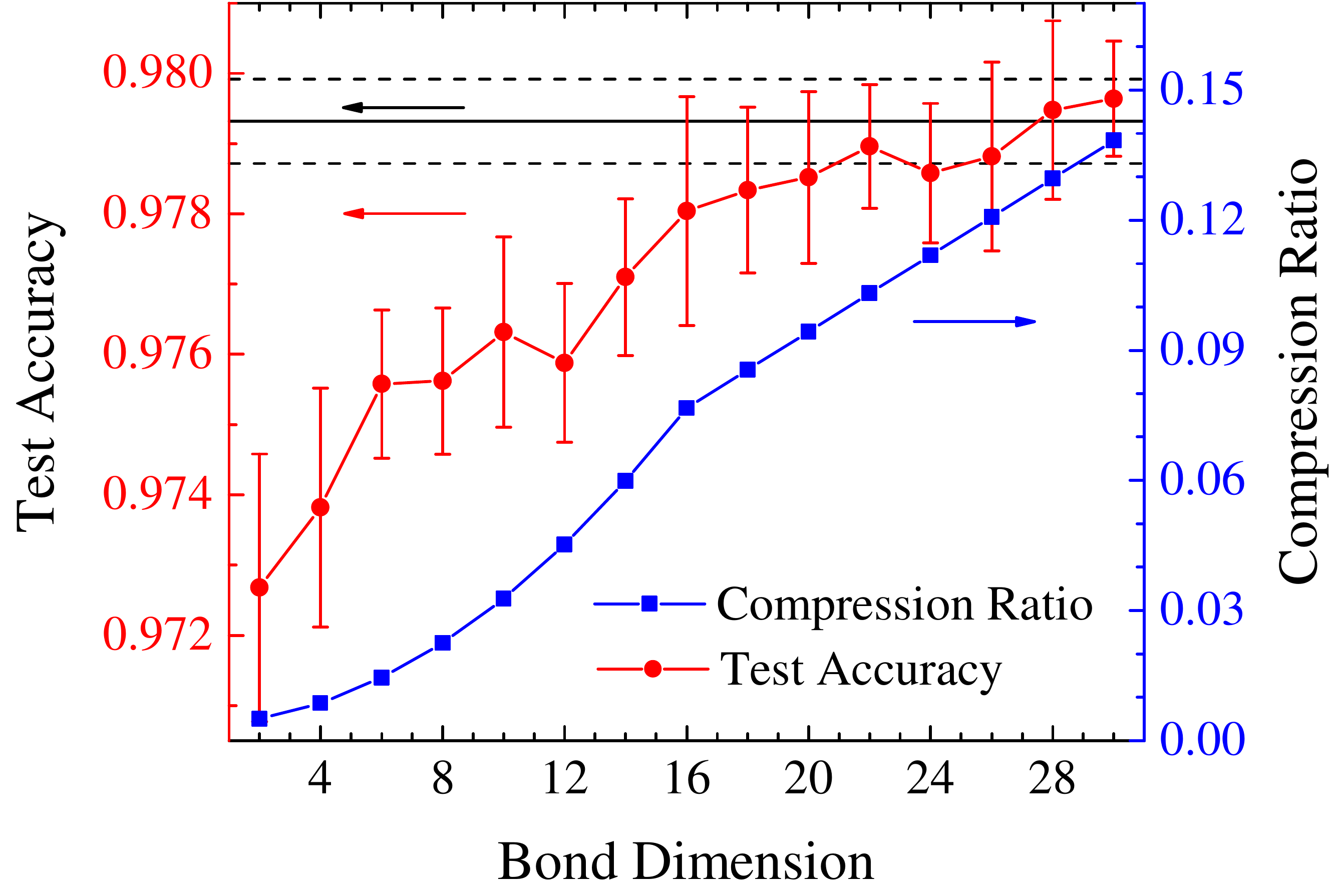}
\caption{Comparison of the test accuracy $a$ between the FC2 and the corresponding MPO-Net on MNIST dataset, without L2 regularization. The corresponding compression ratio $\rho$ is also plotted for each bond dimension $D$. }
\label{Fig:L2norm}
\end{figure}

In Fig.~\ref{Fig:L2norm}, the comparison between FC2 and the corresponding MPO-Net on MNIST is shown. It can be seen clearly that though both the two results are little worse than the results obtained with L2 normalization, the MPO-Net can still reproduce the FC2 results when $D$ becomes large, e.g., $D\geq 16$. Similar conclusion can be drawn from Tab.~\ref{Tab:L2norm}, in which the comparison between VGG networks and the corresponding MPO-Nets on CIFAR-10 is shown, i.e., under such situations, the MPO-Nets can still work at least equally well with normal VGG networks. Thus, we could conclude that the validity of the MPO-Nets is irrelevant to the L2 regularization.

\begin{table}[t]
\centering
\begin{tabular}{c|c|c|c|c}   \hline\hline
     \multirow{2}*{Dataset} &  \multirow{2}*{Network}  & \multicolumn{1}{c|}{Original Rep.} & \multicolumn{2}{c}{MPO-Net}  \\
      \cline{3-5}
      & & $a$ ($\%$) &  $a$ ($\%$) & $\rho$
       \\ \hline \hline
    CIFAR-10 & VGG-16 & 92.39  & 92.90 & 0.0005 \\ \hline   \hline
    CIFAR-10 & VGG-19 & 92.48  & 92.91 & 0.0005 \\ \hline   \hline
\end{tabular}
\caption{Comparison of the test accuracy obtained between VGG networks and the corresponding MPO-Nets. None of the networks here uses L2 regularization.}
\label{Tab:L2norm}
\end{table}

\subsection{Convergence of training}\label{sec:Conv}
One of the motivations of using MPO to represent the linear transformations in deep neural networks is that, with fewer parameters, the training can be expected much easier to converge. However, due to the strong nonlinearity of the cost function, as expressed in Eq.~(7) in the main text, more parameters means higher-dimensional variational space and might have more local minima, thus it is difficult for the current optimization approach, e.g., stochastic gradient descent method, to guarantee a faster convergence speed in a model with fewer parameters. This can be seen from the several networks we used in this paper. E.g., as listed in Tab.~\ref{Tab:NandE}, FC2 has more parameters than LeNet-5, but it converges faster, while the DenseNet has much fewer parameters than VGG-19 and ResNet, however, it needs more epoches to converge the training accuracy.

\begin{table}[t]
\centering
\begin{tabular}{c|c|c|c}   \hline\hline
     DataSet &  Network  & Total parameters & Necessary epoches for training      \\ \hline \hline
  \multirow{2}*{MNIST} &  \multirow{2}*{}  FC2   & 203,264    &   $\sim$ 25  \\   \cline{2-4}
   &   LeNet-5     &    61,470  & $\sim$ 40
      \\  \hline \hline
  \multirow{3}*{CIFAR-10} &  \multirow{3}*{}   VGG-19       & 38,934,208    &   $\sim$ 150    \\   \cline{2-4}
   &   ResNet (k=4, m=18)     &    26,646,448  & $\sim$ 150    \\   \cline{2-4}
   &   DenseNet (n=16, k=m=12)   &    1,001,616 &    $\sim$ 300
      \\  \hline \hline
\end{tabular}
\caption{Extra information about some normal networks used in this work, including the total parameters in the networks and the necessary epoches for the training procedure.}
\label{Tab:NandE}
\end{table}

\begin{figure}[htbp]
\centering
\subfigure[$\chi=2$.]{
\begin{minipage}[t]{0.33\linewidth}
\centering
\includegraphics[width=\linewidth]{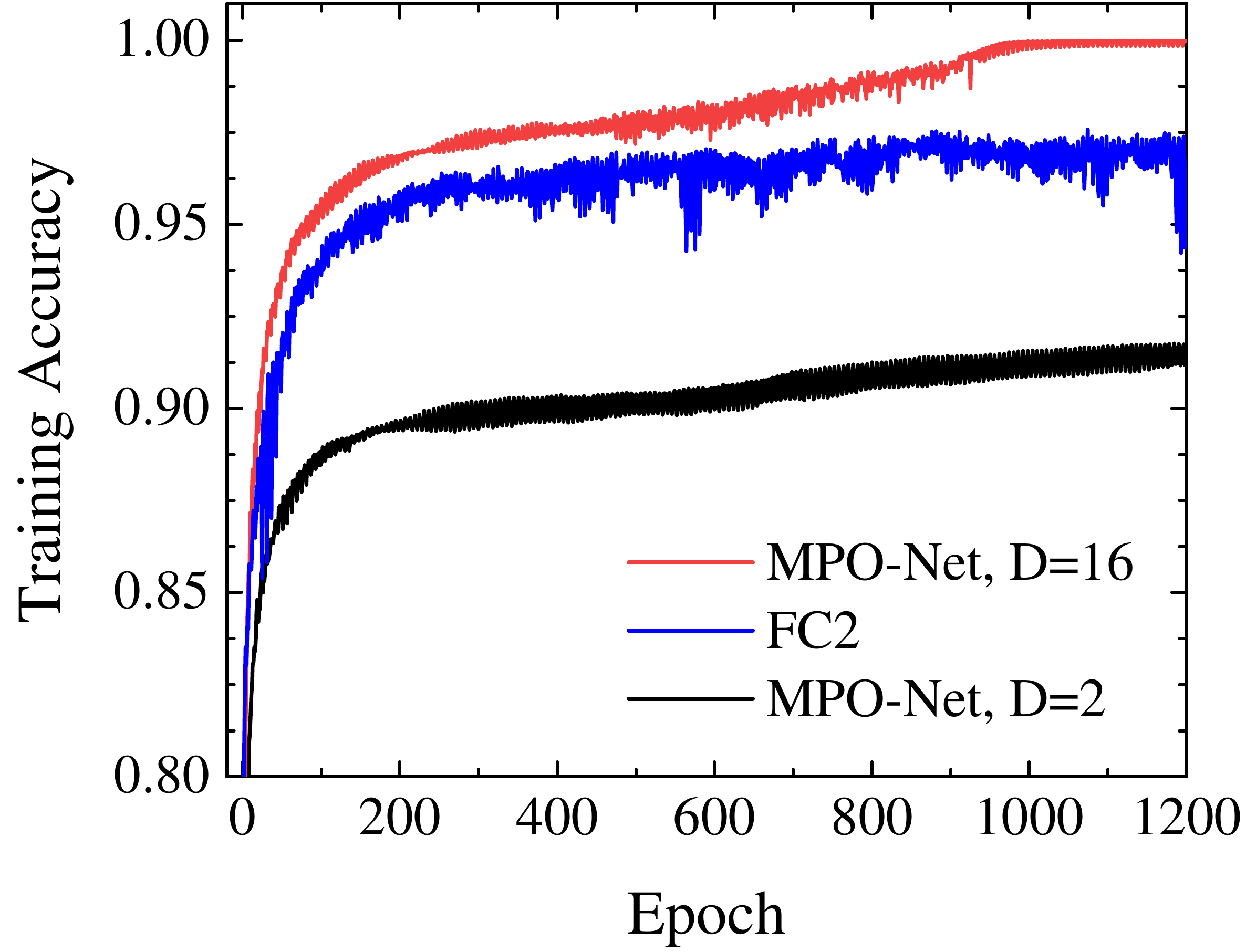}
\end{minipage}%
}%
\subfigure[$\chi=10$.]{
\begin{minipage}[t]{0.33\linewidth}
\centering
\includegraphics[width=\linewidth]{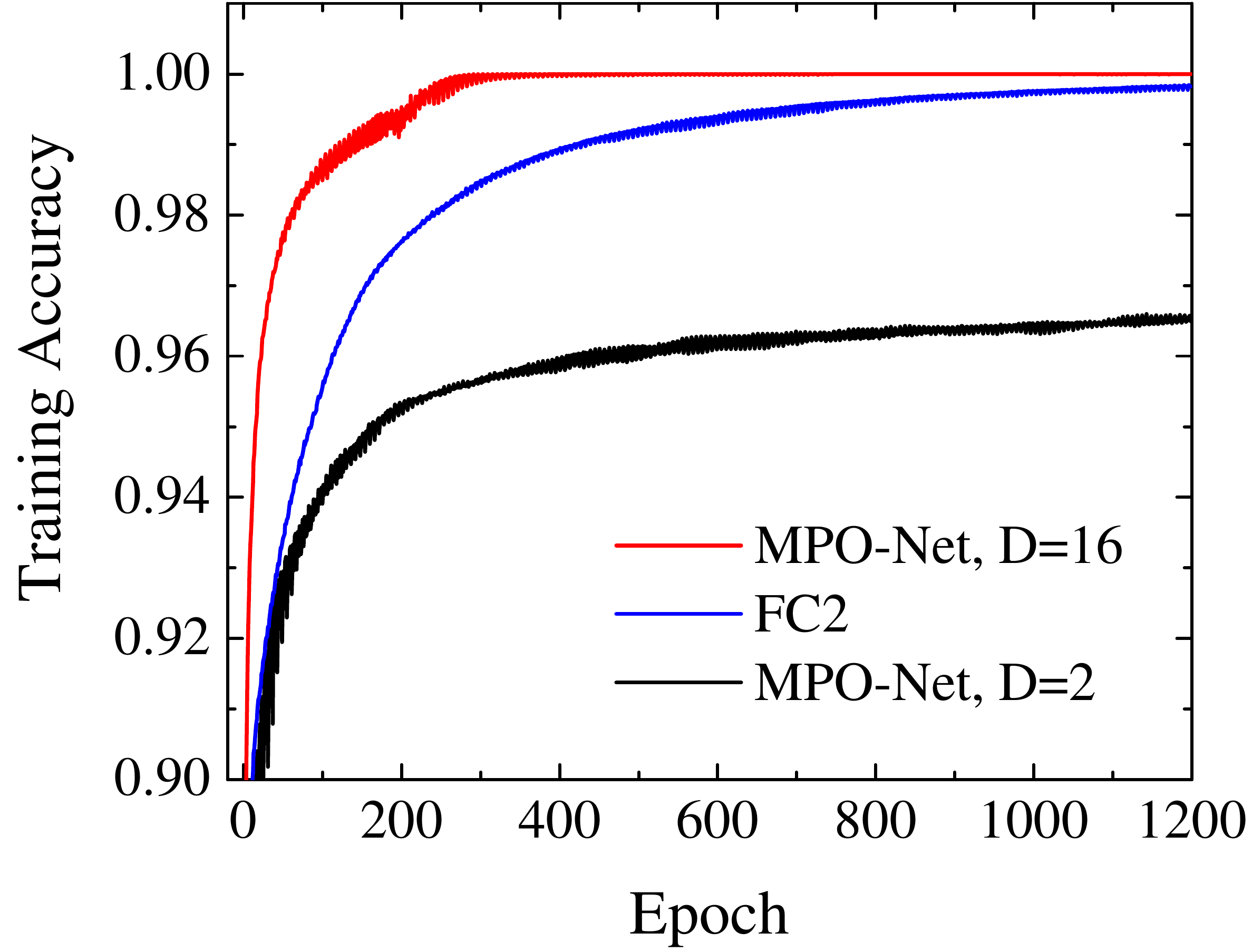}
\end{minipage}%
}%
\subfigure[Orignal data ($\chi = 28$)]{
\begin{minipage}[t]{0.33\linewidth}
\centering
\includegraphics[width=\linewidth]{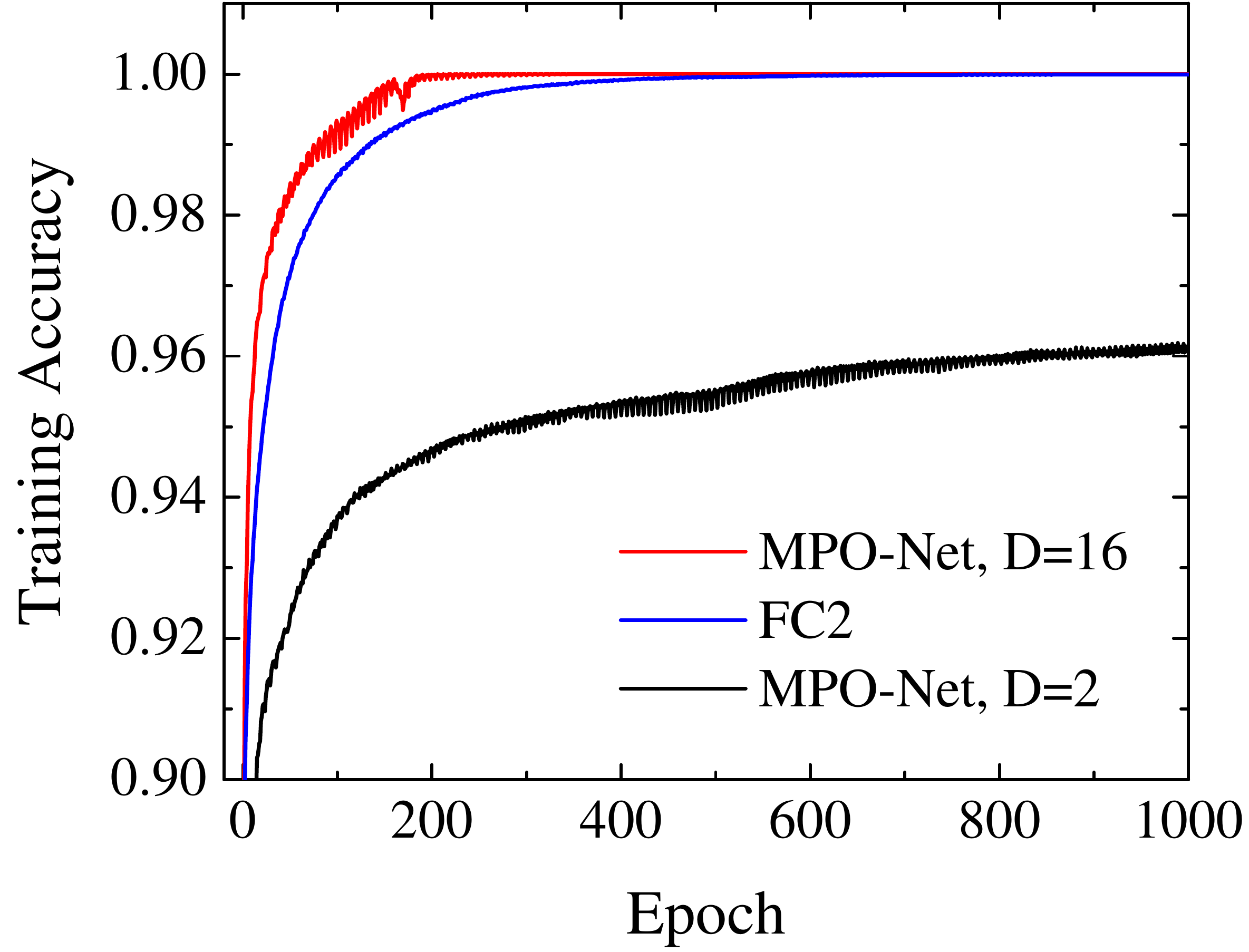}
\end{minipage}
}%
\caption{Comparison of the train accuracy obtained between FC2 and MPO-Nets on MNIST, with different representations of the input data on MNIST. Parameter $\chi$ is used to truncate the MPS and to control the entanglement entropy embodied in the representation.}
\label{Fig:DiffInput}
\end{figure}

Things are more complicated in the case of MPO-Nets. More study shows that it is also related the fact that how much relevant entanglement is hidden in the input data and how much entanglement is grasped by the MPO-Nets. To show this, in Fig.~\ref{Fig:DiffInput}, we plot the training procedures for FC2 and the corresponding MPO-Nets on MNIST dataset. As done in Ref.~\cite{ImagEnt}, similarly we can represent the input images in terms of matrix product states (MPS) \cite{MPS}, and as proposed in the discussion section in the main text, then we can combine the MPS representation of the iamges and the MPO-Nets together to form a complete new network. Suppose the bond dimension of the MPS is denoted as $\chi$, then we know in quantum physics that $\chi$ controls the maximal entanglement grasped in the MPS representation. For MNIST, each image is a $28 \times 28$ square matrix, thus the maximal meaningful $\chi$ is 28, which corresponds to the original image, as shown in (c) in Fig.~\ref{Fig:DiffInput}. We can also throw some irrelevant information by truncating the MPS and use a smaller bond dimension $\chi$ for the input data, as shown in (a) and (b) in Fig.~\ref{Fig:DiffInput}. In all the three cases, we have compared the convergence behavior of the training procedures among two different MPO-Nets and the original FC2 network.

It shows that, when $\chi$ is small, e.g. $\chi = 2$, there is very little entanglement in the input data, the MPO-Net with $D=2$ is underfitting and converges much slowly, while MPO-Net with $D=16$ can produce a higher training accuracy at the same training steps and eventually converges to accuracy $100\%$ after more epoches than FC2. As $\chi$ becomes larger, e.g., $\chi = 10$ and the original data ($\chi = 28$), there is more entanglement represented in the input data, the MPO-Net with $D=2$ is still some kind of underfitting and converges slowly, while MPO-Net with $D=16$ can indeed converge much faster than the original FC2 and obtain a higher accuracy at the same training step. Therefore, as to the MPO-Nets, although in most cases they usually converge faster than the original network, it is difficult to guarantee a faster convergence for the training due to the complicated nonlinear structure of the cost function, the convergence behavior of the optimization methods, as shown in Tab.~\ref{Tab:NandE}, the entanglement entropy hidden in the input data, and the grasped entanglement entropy in the MPO representations, as shown in Fig.~\ref{Fig:DiffInput}. This provides the possibility of analyzing the long-standing unsolved optimization problems in deep learning by entanglement entropy developed in quantum physics.
A complete understanding of the exact relationship can probably improve the current learning algorithm, which will be a pursuit in the future work.

\end{document}